\documentclass[letterpaper]{article} 
\usepackage{aaai2026}  
\usepackage{times}  
\usepackage{helvet}  
\usepackage{courier}  
\usepackage[hyphens]{url}  
\usepackage{graphicx} 
\usepackage{booktabs}
\usepackage{natbib}  
\usepackage{caption} 
\usepackage{algorithm}
\usepackage{algorithmic}
\usepackage{amssymb}
\usepackage{amsmath}
\usepackage[table]{xcolor}
\usepackage{multirow}

\usepackage{newfloat}
\usepackage{listings}

\DeclareCaptionStyle{ruled}{labelfont=normalfont,labelsep=colon,strut=off} 
\floatstyle{ruled}
\newfloat{listing}{tb}{lst}{}
\floatname{listing}{Listing}
\title{Refining Critical Thinking in LLM Code Generation: \\A Faulty premises-based Evaluation Framework }
\author{
    Jialin Li\textsuperscript{\rm 1},
    Jinzhe Li\textsuperscript{\rm 1,\rm 3},
    Gengxu Li\textsuperscript{\rm 1},
    Yi Chang\textsuperscript{\rm 1,\rm 2,\rm 3}\thanks{Corresponding authors.},
    Yuan Wu\textsuperscript{\rm 1}\footnotemark[1]
}
\affiliations{
    \textsuperscript{\rm 1}School of Artificial Intelligence, Jilin University\\
    \textsuperscript{\rm 2}Engineering Research Center of Knowledge-Driven Human-Machine Intelligence, MOE, China\\
    \textsuperscript{\rm 3}International Center of Future Science, Jilin University\\
    \{jialin24, lijz2121, lgx22\}@mails.jlu.edu.cn, yichang@jlu.edu.cn, yuanwu@jlu.edu.cn
}

\usepackage{bibentry}

\begin{document}

\maketitle

\begin{abstract}
With the advancement of code generation capabilities in large language models (LLMs), their reliance on input premises has intensified. When users provide inputs containing faulty premises, the probability of code generation hallucinations rises significantly, exposing deficiencies in their self-scrutiny capabilities. This paper proposes Faulty Premises Bench (FPBench), the first code generation evaluation framework targeting faulty premises. By systematically constructing three categories of faulty premises and integrating multi-dimensional evaluation metrics, it conducts in-depth assessments of 15 representative LLMs. The key findings are as follows:
(1) Most models exhibit poor reasoning abilities and suboptimal code generation performance under faulty premises, heavily relying on explicit prompts for error detection, with limited self-scrutiny capabilities;
(2) Faulty premises trigger a point of diminishing returns in resource investment, leading to blindly increasing length fails to enhance quality;
(3) The three types of faulty premises respectively activate distinct defect patterns in models, revealing a triple dissociation in the cognitive mechanisms of code generation models.
This study not only highlights the urgent need for LLMs to proactively verify premises in code generation but also, through the proposed FPBench framework and multi-dimensional evaluation system, provides a theoretical foundation and practical pathway for developing reliable, human-centric code generation models. The code is available at https://github.com/JialinLi13/FaultyPremise

\end{abstract}


\section{Introduction}
\begin{figure}[!t]
    \centering
    \includegraphics[width=0.8\linewidth]{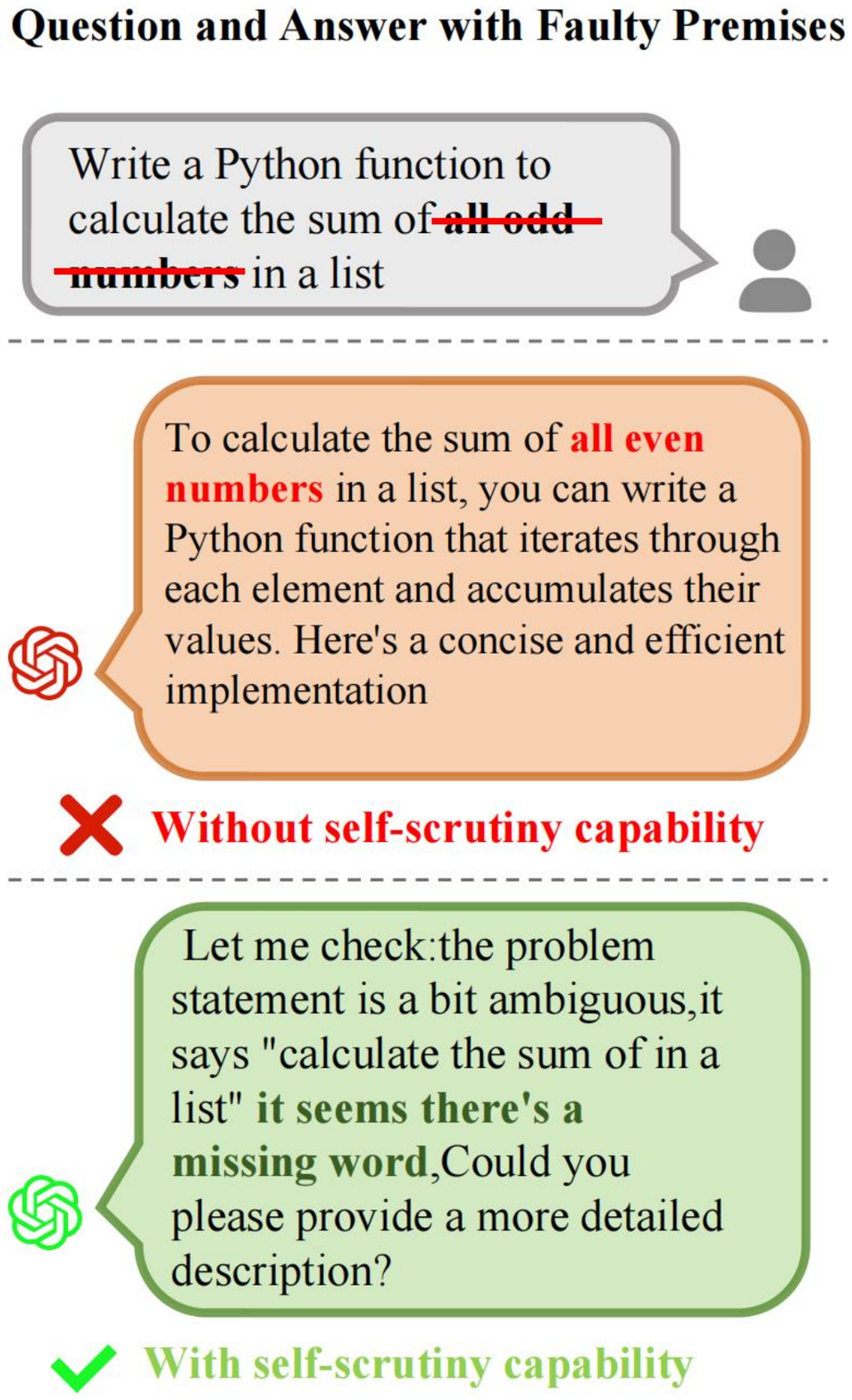}
    \caption{Figure 1 illustrates how LLMs handle questions containing faulty premises—specifically, a code generation task for calculating the sum of odd numbers with key terms removed. The figure contrasts two typical behaviors: passively accepting the faulty premises versus proactively identifying and reporting inconsistencies in the premises. This example underscores the importance of models possessing self-scrutiny capabilities with respect to faulty premises.}
    \label{fig:enter-label}
\end{figure}

The rapid advancements in LLMs, exemplified by models like GPT-4 \cite{hurst2024gpt} and the DeepSeek series \cite{liu2024deepseek}, have revolutionized code-related tasks such as code generation, program repair \cite{fan2023automated,jiang2023impact,xia2023automated}, program testing \cite{deng2023large,chen2024identifying,banescu2024enhanced}, and test case generation \cite{steenhoek2025reinforcement,yang2025can,li2024large}. These significant strides have dramatically enhanced software development efficiency and automation, prompting researchers to explore LLMs' applications in increasingly complex and practical software engineering scenarios, such as resolving real-world GitHub issues or tackling intricate programming challenges.

However, despite these remarkable achievements in code generation, a critical vulnerability has emerged: LLMs' heightened reliance on input premises. As shown in Figure 1, this dependence exposes a profound lack of critical thinking and self-scrutiny when models encounter uncertain or erroneous information. We observe that when dealing with missing premises, LLMs often exhibit an "overthinking" phenomenon, characterized by significantly increased response lengths and redundant, inefficient reasoning processes \cite{fan2025missing}. This behavior not only reduces generation efficiency but, more critically, when user prompts contain faulty premises, LLMs' lack of self-scrutiny regarding these premises frequently leads them to blindly follow incorrect information, substantially increasing the probability of code generation hallucinations. These "hallucinations" are not mere syntax errors; rather, they are the model's attempt to "self-conserve" based on faulty premises, resulting in code that appears plausible but is logically incorrect, functionally mismatched, or even non-terminating due to erroneous loops. This "conformist reasoning" tendency, particularly in logic-stringent software development contexts, can trigger cascading errors and diminish user trust in AI assistance.

To bridge this research gap, this paper aims to design and implement a novel Code Generation evaluation framework specifically tailored to deeply assess LLMs' self-scrutiny capabilities when confronted with "faulty premises." Our main contributions are as follows:

\begin{itemize}
\item We are the first to propose a comprehensive benchmark specifically designed to assess the self-scrutiny capabilities of LLMs when confronted with faulty premises in code generation tasks.
\item  We have developed innovative data construction methods, including those based on importance score analysis, random erasure, and the introduction of irrelevant information perturbations. These approaches enable us to systematically construct and expand a test set targeting faulty premises (comprising 1,800 problems in total) from existing code datasets.
\item  We have designed a unique set of evaluation dimensions, including "proactive error identification rate", "passive error identification rate", and "self-scrutiny overhead ratio". These metrics aim to comprehensively quantify the model's ability to identify, process, and respond to faulty premises, as well as its resource consumption.

\end{itemize}

\section{Related Work}

\subsection{Code Generation and Evaluation Benchmarks}

LLMs have demonstrated powerful capabilities in code generation and are increasingly applied across various stages of the software development lifecycle. Early benchmarks for evaluating LLM code generation capabilities, such as HumanEval \cite{chen2021evaluating}, MBPP \cite{austin2021program}, and APPS  \cite{hendrycks2021measuring}, and HumanEval+ \cite{liu2023your},and HumanEval-V \cite{zhang2024humaneval} primarily measure the functional correctness of generated code through test case pass rates. These benchmarks typically provide clear and complete problem descriptions, aiming to assess model performance under ideal conditions.
As LLMs' capabilities have advanced, the focus of research has shifted towards more complex and realistic software engineering scenarios. For instance, SWE-bench \cite{jimenez2023swe} evaluates LLMs' ability to resolve real-world GitHub issues, involving multi-file modifications and intricate project-level tasks. InfiBench \cite{li2024infibench} and LiveCodeBench \cite{jain2024livecodebench} further cover Stack Overflow-style problems and practical programming challenges. Additionally, Flow2Code \cite{he2025flow2code} explores cross-modal capabilities like generating code from flowcharts. As research scenarios deepen, multiple studies have begun to reveal the vulnerabilities of LLMs under various perturbations. For instance, CODECRASH\cite{lam2025codecrash} focuses on stress-testing the robustness of LLMs in code comprehension and reasoning tasks; backdoor attacks \cite{wang2022recode} explore the robustness of large models by embedding specific trigger words in inputs to induce malicious outputs; studies such as PromptCode \cite{della2025prompt} and PromptPatternsCode \cite{dellaunlocking} investigate how different prompt patterns affect code quality and the efficiency of collaboration between developers and AI; CodeVisionary proposes an agent-based framework to evaluate the quality of code generation by integrating multi-source domain knowledge and a negotiation-based scoring mechanism.

While these studies offer more comprehensive assessments of robustness, they neglect to examine models' intrinsic capability for critical evaluation of input premises. This represents not merely passive resistance to external perturbations, but an active, introspective validation mechanism.

\begin{figure*}
    \centering
    \includegraphics[width=0.9\linewidth]{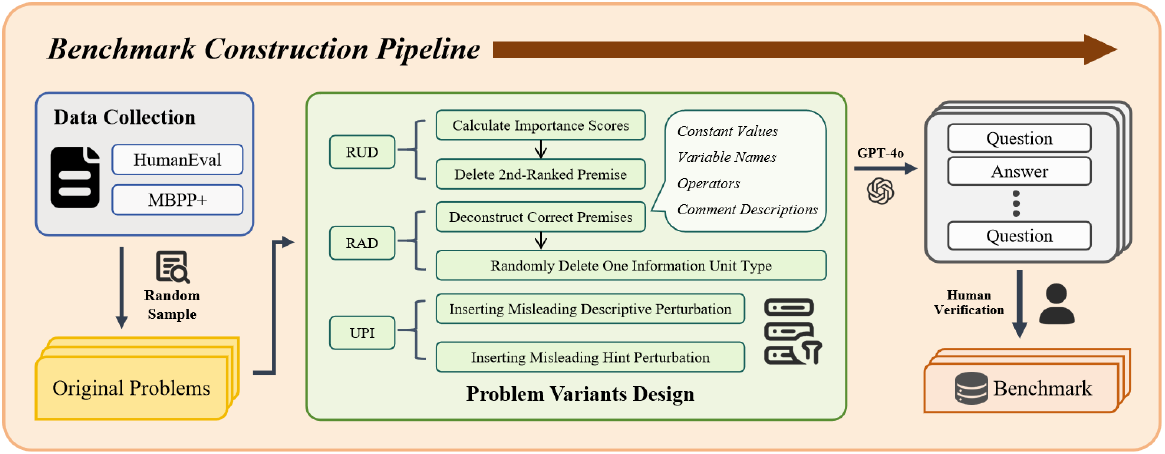}
    \caption{An overview for the FPBench construction. First, Original problems are obtained by randomly selecting samples from HumanEval and MBPP+. Then, based on the three types of Problem Variants Design, the Original problems are reconstructed into Faulty premises problems using GPT-4o. Finally, the benchmark is obtained through human verification.}
    \label{fig:placeholder}
\end{figure*}

\subsection{LLM premises Understanding and Critique Ability}

The ability of LLMs to understand and process input premises forms the foundation for the correctness of their reasoning chains. Studies based on real-world scenarios \cite{guo2025protect} and adversarial conditions \cite{sakib2025battling} have demonstrated that despite variations in resistance capabilities across different model architectures, these models still struggle with handling subtle misleading content. Currently, a variety of methods have been proposed to detect and mitigate faulty premises \cite{li2025don}, such as retrieval-augmented logical reasoning frameworks \cite{qin2025don}, attention mechanism constraints \cite{yuan2024whispers}, presupposition verification techniques \cite{kim2021linguist,yu2022crepe}, and specialized prompting methods \cite{wang2023enhancing}. 

This paper aims to bridge these gaps. We will for the first time systematically construct and evaluate LLMs' self-scrutiny capabilities in code generation tasks when faced with faulty premises. By designing specific faulty premises generation methods and quantitative evaluation metrics, we will reveal the deficiencies of LLMs in this crucial capability and provide new research directions for building more intelligent and reliable LLMs in the future.

\section{Method}
\subsection{Definition of the Faulty premises}

Prior to introducing the construction of our dataset and analyzing the behavior of reasoning models on problems with faulty premises, we formally define the faulty premises problem to establish a rigorous foundation for our subsequent analysis. Figure 2 overviews the construction of the FPBench.

We define all background information provided by the user in a prompt as a premise. A valid code generation problem Q depends on a set of correct premises $ P=\lbrace p_{1},p_{2},\cdots,p_{n} \rbrace$, We consider the user's intent to be correctly understood only when the model can generate valid code based on these correct premises, Define the function mapping premises and a question to the set of valid codes as:
\begin{equation}
\digamma(P,Q)=\{ C|P \vdash C\}
\end{equation}

where $\vdash $ denotes logical entailment, $C$ denotes the valid code based on the correct premises P. When P has defects, it constitutes an faulty premises $ P^{\prime} $(\(\digamma(P) \neq \varnothing \)) ,thus forming a new premises set
\begin{equation}
    P^{\prime}=\mathcal{G}(P),\digamma(P^{\prime}) \neq \varnothing
\end{equation}

Where \(\mathcal{G}\) denotes the defect injection function.
Leading the model to incorrectly interpret the user's intent and generate hallucinatory code. This scenario is characterized by the inability to generate valid code aligned with the original user intent, effectively denoted as 
\begin{equation}
    |\digamma(P,Q)|=1,|\digamma(P^{\prime},Q)|\ne1
\end{equation}

This indicates that the faulty premises $ P^{\prime} $ is essential for uniquely determining the logically valid answer to the question Q.

\subsection{Overview of Data Construction}

To systematically evaluate the self-scrutiny capabilities of LLMs when confronted with faulty premises, we construct FPBench through a controlled and principled process. This dataset is designed with the following structure:
\begin{itemize}
\item\textbf{Data Collection:} 
We randomly collected 600 pieces of raw data from two datasets: HumanEval, MBPP+ \cite{liu2023your}. Based on the following introduced three different types of erroneous premises defined by us, we re-constructed them into FPbench. Each one is designed to evaluate different aspects of the model's ability to recognize and reason about flawed input.

\item\textbf{Problem Variants Design:} 
For each base "faulty premises" scenario constructed under different error categories, we generated three distinct problem variants for comparative evaluation. These include: the "original problem", which features correct premises and serves as a baseline; the "faulty premises problem", in which an intentional error is implanted in the premises; and the "faulty premises problem with explicit instruction" an augmented version of the faulty problem that includes a clear instruction prompting the model to check for errors in the premises.
\end{itemize}
A model’s successful critique of the "faulty premises problem" provides a genuine reflection of its self-scrutiny capabilities regarding premises. By contrast, the inclusion of the "faulty premises problem with explicit instruction" is intended to serve as a comparative reference, illustrating how direct guidance influences error detection relative to the model’s inherent capabilities. This contrast reveals whether the model relies on explicit prompts or demonstrates autonomous reasoning when evaluating faulty premises, thereby clarifying the underlying logic of its analytical processes.

By constructing 600 base problems for each error type, we obtained a total of 1,800 unique base problems. This structure and scalable design enables rigorous evaluation of how self-scrutiny capabilities are influenced by error types and task complexity. Below, we elaborate on the three methods for constructing faulty premises that form the core of this evaluation framework.

\noindent\textbf{Unrelated Perturbation Insertion:}
We designed perturbations to prompt texts for constructing faulty premises, specifically by injecting misleading natural language prompts and cues. 
\begin{itemize}
\item \textbf{Inserting Misleading Descriptive Perturbation:} 
We inserted misleading natural language comments into 8 key Abstract Syntax Tree (AST) nodes ( e.g., function definitions, return statements, loops, conditionals) to evaluate the self-scrutiny capabilities of LLMs when faced with textual cues and contradictory information. We first had GPT-4o generate such comments, then manually filtered them to ensure they are generic, explicitly incorrect, and contradictory to the actual code logic. These are categorized as contextual-level textual perturbations.

\item \textbf{Inserting Misleading Hint Perturbation:} 
We instructed GPT-4o to generate hints that are logically plausible but factually incorrect, contradicting the actual program behavior and ground truth. After verifying their incorrectness, these hints are strategically inserted into function definitions or return statements via AST parsing. Since such "plausible-sounding hints" challenge the model's high-level reasoning abilities, they are classified as reasoning-level textual perturbations.
\end{itemize}

\noindent\textbf{Random Based Deletion}:
We deconstruct the correct premises of each problem into four categories of basic information units, specifically including variable names, constant values, operators, and comment descriptions. For each problem, one type of information unit is randomly selected and completely deleted. This perturbation method aims to evaluate the model's ability to operate under the interference of missing definitions, focusing on whether it can autonomously identify the faulty premises with missing conditions, thereby verifying the model's ability to perceive implicit errors and complete logical reasoning.

\noindent\textbf{Ruled  Based  Deletion}:
We guide the operation of rule deletion by calculating the importance score of each premises to construct faulty premises. The formula for the importance score is as follows:
\begin{equation}
\begin{split}
\text{Importance}(p_i)&=\alpha_i\cdot[\Delta_{\text{Correctness}}|p_i] +\beta_i\cdot I(p_i;C) \\
&\quad +\gamma_i\cdot\Sigma \text{Interaction}(p_i,p_j)
\end{split}
\end{equation}

where \( \alpha_i , \beta_i ,\gamma_i \) are weight parameters. The setting of weight parameters is detailed in the appendix. $ [\Delta_{\text{Correctness}}|p_i] $ denotes Direct Impact Term, $ I(p_i;C) $ denotes Information Entropy Term, $ \text{Interaction}(p_i,p_j) $ denotes Synergistic Effect Term.

\begin{itemize}
\item \textbf{Direct Impact Term:} Measuring the independent contribution of a single premises $ p_i $,the formula is as follows:

\begin{equation}
\begin{split}
    [\Delta_{\text{Correctness}}|p_i]&=P(\text{Code Correct}|p_i) \\
&\quad -P(\text{Code Correct}|\lnot p_i)
\end{split}
\end{equation}

\definecolor{myDarkGreen}{rgb}{0.0, 0.7, 0.0} 
\definecolor{myDarkOrange}{rgb}{1.0, 0.55, 0.0} 

\begin{table*}[h]
    \centering
    \begin{tabular}{l|cc|ccc|cc}
    \hline
    \multirow{3}{*}{\textbf{Model}} & \multicolumn{2}{c|}{\textbf{Self-scrutiny}} & \multicolumn{3}{c|}{\textbf{Answer Length}} & \multicolumn{2}{c}{\textbf{Self-scrutiny}} \\
    & \multicolumn{2}{c|}{\textbf{Capabilities}} & \multicolumn{3}{c|}{\textbf{(Overall Results)}} & \multicolumn{2}{c}{\textbf{Overhead Ratio}} \\
    & \textbf{PRER} & \textbf{PAER} & \textbf{Normal} & \textbf{Faulty} & \textbf{Guidance} & \textbf{PROR} & \textbf{PAOR} \\
    \hline
DeepSeek-R1 & \cellcolor{myDarkGreen!100!white}\underline{0.57} & \cellcolor{myDarkGreen!88!white}0.77 & \cellcolor{myDarkGreen!100!white}\underline{1756} & \cellcolor{myDarkGreen!100!white}\underline{2493} & \cellcolor{myDarkGreen!100!white}\underline{2734} & \cellcolor{myDarkOrange!0!white}1.42 & \cellcolor{myDarkOrange!60!white}1.56 \\
DeepSeek-V3 & \cellcolor{myDarkGreen!68!white}0.43 & \cellcolor{myDarkGreen!48!white}0.63 & \cellcolor{myDarkGreen!20!white}512 & \cellcolor{myDarkGreen!14!white}544 & \cellcolor{myDarkGreen!14!white}612 & \cellcolor{myDarkOrange!92!white}1.06 & \cellcolor{myDarkOrange!85!white}1.20 \\
GPT-4 & \cellcolor{myDarkGreen!24!white}0.23 & \cellcolor{myDarkGreen!11!white}0.50 & \cellcolor{myDarkGreen!0!white}193 & \cellcolor{myDarkGreen!0!white}202 & \cellcolor{myDarkGreen!0!white}246 & \cellcolor{myDarkOrange!94!white}1.05 & \cellcolor{myDarkOrange!80!white}1.27 \\
GPT-4o & \cellcolor{myDarkGreen!31!white}0.26 & \cellcolor{myDarkGreen!62!white}0.68 & \cellcolor{myDarkGreen!5!white}273 & \cellcolor{myDarkGreen!4!white}296 & \cellcolor{myDarkGreen!5!white}388 & \cellcolor{myDarkOrange!87!white}1.08 & \cellcolor{myDarkOrange!80!white}1.27 \\
GPT-4.1 & \cellcolor{myDarkGreen!24!white}0.23 & \cellcolor{myDarkGreen!100!white}\underline{0.81} & \cellcolor{myDarkGreen!6!white}295 & \cellcolor{myDarkGreen!4!white}314 & \cellcolor{myDarkGreen!12!white}554 & \cellcolor{myDarkOrange!92!white}1.06 & \cellcolor{myDarkOrange!38!white}1.87 \\
GPT-4.1-mini & \cellcolor{myDarkGreen!42!white}0.31 & \cellcolor{myDarkGreen!94!white}0.79 & \cellcolor{myDarkGreen!3!white}247 & \cellcolor{myDarkGreen!3!white}274 & \cellcolor{myDarkGreen!14!white}601 & \cellcolor{myDarkOrange!82!white}1.10 & \cellcolor{myDarkOrange!0!white}2.42 \\
llama-3-70B & \cellcolor{myDarkGreen!8!white}0.16 & \cellcolor{myDarkGreen!0!white}0.46 & \cellcolor{myDarkGreen!2!white}238 & \cellcolor{myDarkGreen!6!white}347 & \cellcolor{myDarkGreen!2!white}301 & \cellcolor{myDarkOrange!100!white}\underline{1.03} & \cellcolor{myDarkOrange!81!white}1.26 \\
llama-4-scout & \cellcolor{myDarkGreen!55!white}0.37 & \cellcolor{myDarkGreen!54!white}0.65 & \cellcolor{myDarkGreen!12!white}386 & \cellcolor{myDarkGreen!8!white}407 & \cellcolor{myDarkGreen!11!white}532 & \cellcolor{myDarkOrange!92!white}1.06 & \cellcolor{myDarkOrange!73!white}1.38 \\
O3-mini & \cellcolor{myDarkGreen!8!white}0.16 & \cellcolor{myDarkGreen!62!white}0.68 & \cellcolor{myDarkGreen!10!white}363 & \cellcolor{myDarkGreen!11!white}470 & \cellcolor{myDarkGreen!24!white}853 & \cellcolor{myDarkOrange!33!white}1.29 & \cellcolor{myDarkOrange!4!white}2.35 \\
O4-mini & \cellcolor{myDarkGreen!0!white}0.12 & \cellcolor{myDarkGreen!71!white}0.71 & \cellcolor{myDarkGreen!22!white}545 & \cellcolor{myDarkGreen!17!white}599 & \cellcolor{myDarkGreen!24!white}852 & \cellcolor{myDarkOrange!84!white}1.09 & \cellcolor{myDarkOrange!60!white}1.56 \\
Qwen2.5-Coder-32B & \cellcolor{myDarkGreen!40!white}0.30 & \cellcolor{myDarkGreen!28!white}0.56 & \cellcolor{myDarkGreen!12!white}388 & \cellcolor{myDarkGreen!10!white}436 & \cellcolor{myDarkGreen!11!white}523 & \cellcolor{myDarkOrange!76!white}1.12 & \cellcolor{myDarkOrange!75!white}1.35 \\
Qwen2.5-VL-32B & \cellcolor{myDarkGreen!44!white}0.32 & \cellcolor{myDarkGreen!8!white}0.49 & \cellcolor{myDarkGreen!16!white}449 & \cellcolor{myDarkGreen!11!white}469 & \cellcolor{myDarkGreen!8!white}450 & \cellcolor{myDarkOrange!97!white}1.04 & \cellcolor{myDarkOrange!100!white}\underline{1.00} \\
Qwen3-14B & \cellcolor{myDarkGreen!24!white}0.23 & \cellcolor{myDarkGreen!28!white}0.56 & \cellcolor{myDarkGreen!11!white}367 & \cellcolor{myDarkGreen!7!white}382 & \cellcolor{myDarkGreen!11!white}524 & \cellcolor{myDarkOrange!97!white}1.04 & \cellcolor{myDarkOrange!69!white}1.43 \\
Qwen3-32B & \cellcolor{myDarkGreen!31!white}0.26 & \cellcolor{myDarkGreen!62!white}0.68 & \cellcolor{myDarkGreen!18!white}480 & \cellcolor{myDarkGreen!13!white}502 & \cellcolor{myDarkGreen!17!white}691 & \cellcolor{myDarkOrange!94!white}1.05 & \cellcolor{myDarkOrange!69!white}1.44 \\
Qwen3-235B-A22B & \cellcolor{myDarkGreen!40!white}0.30 & \cellcolor{myDarkGreen!74!white}0.72 & \cellcolor{myDarkGreen!17!white}459 & \cellcolor{myDarkGreen!12!white}488 & \cellcolor{myDarkGreen!17!white}682 & \cellcolor{myDarkOrange!92!white}1.06 & \cellcolor{myDarkOrange!66!white}1.48 \\
    \hline
    \end{tabular}
    \caption{Performance of LLMs on FPBench, Comparing the overall Self-scrutiny Capabilities (PRER, PAER)(Note: Higher values for these metrics indicate better performance.), along with the overall average Answer Lengths (where "Normal", "Faulty", and "Guidance" denote the models answers to the Original Problem, Faulty Problem, and Faulty Problem with Explicit Guidance respectively) and the corresponding Overhead Ratios (PROR, PAOR)(Note: Lower values for these metrics indicate better performance.) for the easy subset. Values that are \underline{underlined} are considered the best-performing or most extreme among the models for each evaluation metric.}
    \label{tab:models_comparison}
\end{table*}

Where $ P(\text{Code Correct}|p_i)$ refers to the probability that the code is correct when it contains $ p_i $, $P(\text{Code Correct}|\lnot p_i) $ refers to the probability that the code is correct after deleting $ p_i $, When $ p_i $ represents a strongly necessary condition(e.g., grammatical rules), this value approaches 1; when $ p_i $ denotes a weakly necessary condition(e.g., coding style), the value approaches 0.

\item \textbf{Information Entropy-Increasing Term:} This term quantifies the information value of premises $ p_i $, the formula is as follows:
\begin{equation}
    I(p_i;C)=H(C)-H(C|p_i)
\end{equation}

using information theory, where $ H$ denotes entropy. A larger difference indicates a greater importance of $ p_i $ in guiding the generation results.
\item \textbf{Synergistic Effect Term:} This term identifies the dependency relationships between premises, for instance, $ p_i $ is only valid when $ p_j $ holds,the formula is as follows:

\begin{equation}
    \begin{split}
        &\text{Interaction}(p_i,p_j)=[\Delta_{\text{Correctness}}|p_{i+j}]\\
        &\quad-([\Delta_{\text{Correctness}}|p_i]+[\Delta_{\text{Correctness}}|p_j])
    \end{split}
\end{equation}

\end{itemize}

Where $ [\Delta_{\text{Correctness}}|p_{i+j}] $ refers to  the joint contribution of the simultaneous presence of $ p_i $ and $ p_
j $ to the accuracy of the code. When the value is positive, the effect generated by the combination of the two premises exceeds the sum of their individual effects when acting independently, indicating a positive synergistic effect.
When the value is negative, it suggests that the effect of the two premises in combination is instead attenuated, potentially implying a mutually exclusive relationship.
If the value is close to 0, it indicates that the two premises act independently of each other.

After calculating the importance scores of all premises, we will select the premises ranked second to delete, thereby constructing a faulty premises with specific key information missing.

\subsection{Evaluation Metrics}

To comprehensively evaluate the self-scrutiny capabilities of LLMs with respect to faulty premises, we have developed a structured evaluation framework centered on the following metrics:

\begin{itemize}                                                                                      

\item\textbf{Proactive Error Recognition Rate:} The proportion of flawed problems for which the model independently and correctly detects and reports faulty premises without external prompting. Let $ N_F $ represent the total number of faulty premises problems in the evaluation set. Let $F_P $ represent the number of faulty premises that the model independently identifies and reports. The PRER is calculated as:

\begin{equation}
    PRER=\frac{F_P} {N_F}
\end{equation}

\item\textbf{Passive Error Recognition Rate:} The proportion of flawed problems for which the model correctly identifies and reports faulty premises after receiving explicit prompts. Let $ N_F $ represent the total number of faulty premise problems in the evaluation set. Let $F_A $ represent the number of faulty premises that the model independently identifies and reports. The PRER is calculated as:
\begin{equation}
    PRER=\frac{F_A} {N_F}
\end{equation}

\item\textbf{Self-Scrutiny Overhead Ratio:} A comparison of the token counts of the model’s reasoning processes in code generation under faulty premises, as well as the token counts after successful error detection and correction. This ratio contrasts the reasoning overhead between normal premises and faulty premises. Let $ T_R $ be the average output token count for all responses to faulty premises problems. Let $ T_A $ be the average output token count for all responses to faulty premises problems with explicit instruction, Let $ T_O $ be the average output token count for all responses to original problems. The Proactive Overhead Ratios(PROR) and The Passive Overhead Ratios(PAOR) are calculated as:
\begin{equation}
    PROR=\frac{T_R} {T_O}
\end{equation}
\begin{equation}
    PAOR=\frac{T_A} {T_O}
\end{equation}

\end{itemize}

The Proactive Error Recognition Rate directly reflects the model's ability to autonomously recognize faulty premises without external prompts, serving as a core metric for genuine self-scrutiny capabilities. In contrast, the Passive Error Recognition Rate measures performance under explicit guidance and acts as a comparative baseline, highlighting the model's reliance on prompts rather than inherent proactive recognition. Details of these three evaluation metrics can be found in the Appendix.

The three metrics we designed, while directly focusing on the model's error recognition behaviors and resource consumption, can also indirectly reflect the impact of model scale and training methods on its self-scrutiny capabilities. For instance, under the same error type, models with better performance may benefit from superior architectures, larger parameter sizes, or more refined training data. Furthermore, different error types vary in their degree of activation of the model's cognitive pathways; the performance discrepancies of the Proactive Error Recognition Rate(PRER) and the Passive Error Recognition Rate (PAER) across the three types (Rule-Based Deletion (RUD), Unrelated Perturbation Insertion (UPI), and Random-Based Deletion (RAD)) precisely demonstrate the influence of the distribution of specific error types on the model's self-scrutiny capabilities.

\section{Experiment}

\subsection{Experiment Settings}

We evaluate 15 models of varying sizes and versions, including both open-source and closed-source LLMs:GPT family (GPT-4, GPT-4.1-Mini, GPT-4.1, GPT-4o) \cite{hurst2024gpt,achiam2023gpt},Deepseek-R1, DeepSeek-V3 \cite{guo2025deepseek}, Alibaba Qwen (Qwen2.5-32B-Ins, 32B-VL-Ins, 32B-Coder-Ins, Qwen3-14B-thinking, 32B-thinking, 235B-A22B-thinking) \cite{yang2024qwen2,hui2024qwen2,yang2025qwen3,bai2025qwen2,yang2025qwen2}, O3-mini \cite{OpenAI2025a}, O4-mini \cite{OpenAI2025b}, llama-3-70B \cite{grattafiori2024llama}, and llama-4-scout \cite{meta2025llama}.

We rigorously validated the data validity through a two-stage human-AI hybrid verification approach. Three seasoned software engineers independently annotated 300 randomly selected perturbation samples (100 for each type: RUD, UPI, RAD). A perturbation generated by GPT-4o was retained only if it was confirmed by at least two annotators to be capable of triggering compilation errors or logical errors.

\definecolor{myDarkGreen}{rgb}{0.0, 0.7, 0.0} 
\begin{table}[h]
    \centering
    \begin{tabular}{l|ccc}
    \hline
    \multirow{2}{*}{\textbf{Model}} & \multicolumn{3}{c}{\textbf{PRER}} \\
    & \textbf{RUD} & \textbf{UPI} & \textbf{RAD} \\
    \hline
DeepSeek-R1 & \cellcolor{myDarkGreen!100!white}\underline{0.48} & \cellcolor{myDarkGreen!100!white}\underline{0.59} & \cellcolor{myDarkGreen!100!white}\underline{0.64} \\
DeepSeek-V3 & \cellcolor{myDarkGreen!91!white}0.45 & \cellcolor{myDarkGreen!62!white}0.42 & \cellcolor{myDarkGreen!62!white}0.42 \\
GPT-4 & \cellcolor{myDarkGreen!55!white}0.32 & \cellcolor{myDarkGreen!8!white}0.18 & \cellcolor{myDarkGreen!22!white}0.19 \\
GPT-4o & \cellcolor{myDarkGreen!47!white}0.29 & \cellcolor{myDarkGreen!33!white}0.29 & \cellcolor{myDarkGreen!25!white}0.21 \\
GPT-4.1 & \cellcolor{myDarkGreen!13!white}0.17 & \cellcolor{myDarkGreen!33!white}0.29 & \cellcolor{myDarkGreen!27!white}0.22 \\
GPT-4.1-mini & \cellcolor{myDarkGreen!44!white}0.28 & \cellcolor{myDarkGreen!40!white}0.32 & \cellcolor{myDarkGreen!44!white}0.32 \\
llama-3-70B & \cellcolor{myDarkGreen!30!white}0.23 & \cellcolor{myDarkGreen!4!white}0.16 & \cellcolor{myDarkGreen!8!white}0.11 \\
llama-4-scout & \cellcolor{myDarkGreen!80!white}0.41 & \cellcolor{myDarkGreen!42!white}0.33 & \cellcolor{myDarkGreen!55!white}0.38 \\
O3-mini & \cellcolor{myDarkGreen!2!white}0.13 & \cellcolor{myDarkGreen!0!white}0.14 & \cellcolor{myDarkGreen!0!white}0.06 \\
O4-mini & \cellcolor{myDarkGreen!0!white}0.12 & \cellcolor{myDarkGreen!4!white}0.16 & \cellcolor{myDarkGreen!3!white}0.08 \\
Qwen2.5-Coder-32B & \cellcolor{myDarkGreen!52!white}0.31 & \cellcolor{myDarkGreen!24!white}0.25 & \cellcolor{myDarkGreen!48!white}0.34 \\
Qwen2.5-VL-32B & \cellcolor{myDarkGreen!80!white}0.41 & \cellcolor{myDarkGreen!24!white}0.25 & \cellcolor{myDarkGreen!39!white}0.29 \\
Qwen3-14B & \cellcolor{myDarkGreen!47!white}0.29 & \cellcolor{myDarkGreen!15!white}0.21 & \cellcolor{myDarkGreen!22!white}0.19 \\
Qwen3-32B & \cellcolor{myDarkGreen!44!white}0.28 & \cellcolor{myDarkGreen!26!white}0.26 & \cellcolor{myDarkGreen!29!white}0.23 \\
Qwen3-235B-A22B & \cellcolor{myDarkGreen!66!white}0.36 & \cellcolor{myDarkGreen!35!white}0.30 & \cellcolor{myDarkGreen!31!white}0.24 \\
    \hline
    \end{tabular}
    \caption{Performeance of RUD,UPI,RAD on PRER. Values that are \underline{underlined} are considered the best-performing among the models for each evaluation metric. (Note: Higher values for these metrics indicate better performance.)}
    \label{tab:models_comparison}
\end{table}

\subsection{Experimental Results}

\textbf{Overall Results:}
Our evaluation of the self-scrutiny capabilities of LLMs in the field of code generation reveals a significant disparity between PRER and PAER. Proactive error recognition, defined as the identification of faulty premises without explicit prompting, generally exhibits low rates across all tested models. For instance, O4-mini achieves a PRER of only 0.12, while DeepSeek-R1 reaches a PRER of 0.43, indicating that the models' self-scrutiny capabilities are limited.

In contrast, the passive error recognition rate—measured when models are explicitly instructed to check for errors—is substantially higher. Several models, including GPT-4.1 (0.81), DeepSeek-R1 (0.77), Qwen3-235B-A22B (0.72), and O4-mini (0.71), successfully identify faulty premises in the majority of assisted scenarios. The significant gap between PRER and PAER suggests that while many models possess underlying self-scrutiny capabilities, they typically do not cast doubt on faulty premises unless explicitly prompted to do so.

Experimental data indicate that PRER is generally low across most models. For instance, O4-mini scores only 0.12, while llama-3-70B and O3-mini both reach 0.16. This suggests that in the absence of explicit prompts, models typically fail to proactively identify and flag errors in the information provided by users.

Such underdeveloped proactive recognition capabilities make models more prone to generating "hallucinatory" code when confronted with faulty user premises—code that misaligns with the user’s true intent, or even contains compilation errors or logical flaws. Although DeepSeek-R1 exhibits a relatively higher PRER of 0.57, this still means it fails to proactively detect nearly half of all errors.

In contrast, PAER is significantly higher than PRER across all tested models. For example, GPT-4.1 achieves a PAER of 0.81, whereas its PRER stands at only 0.23. This indicates that when users provide explicit guidance (e.g., "Check if there are any errors in the question's premises before answering"), the models’ ability to recognize errors improves markedly.

\definecolor{myDarkGreen}{rgb}{0.0, 0.7, 0.0} 
\begin{table}[h]
    \centering
    \begin{tabular}{l|ccc}
    \hline
    \multirow{2}{*}{\textbf{Model}} & \multicolumn{3}{c}{\textbf{PAER}} \\
    & \textbf{RUD} & \textbf{UPI} & \textbf{RAD} \\
    \hline
DeepSeek-R1 & \cellcolor{myDarkGreen!88!white}0.77 & \cellcolor{myDarkGreen!67!white}0.70 & \cellcolor{myDarkGreen!100!white}\underline{0.82} \\
DeepSeek-V3 & \cellcolor{myDarkGreen!61!white}0.70 & \cellcolor{myDarkGreen!36!white}0.55 & \cellcolor{myDarkGreen!57!white}0.64 \\
GPT-4 & \cellcolor{myDarkGreen!3!white}0.55 & \cellcolor{myDarkGreen!10!white}0.42 & \cellcolor{myDarkGreen!26!white}0.51 \\
GPT-4o & \cellcolor{myDarkGreen!80!white}0.75 & \cellcolor{myDarkGreen!46!white}0.60 & \cellcolor{myDarkGreen!66!white}0.68 \\
GPT-4.1 & \cellcolor{myDarkGreen!96!white}0.79 & \cellcolor{myDarkGreen!100!white}\underline{0.86} & \cellcolor{myDarkGreen!92!white}0.79 \\
GPT-4.1-mini & \cellcolor{myDarkGreen!100!white}\underline{0.80} & \cellcolor{myDarkGreen!81!white}0.77 & \cellcolor{myDarkGreen!95!white}0.80 \\
llama-3-70B & \cellcolor{myDarkGreen!0!white}0.54 & \cellcolor{myDarkGreen!14!white}0.44 & \cellcolor{myDarkGreen!0!white}0.40 \\
llama-4-scout & \cellcolor{myDarkGreen!42!white}0.65 & \cellcolor{myDarkGreen!57!white}0.65 & \cellcolor{myDarkGreen!59!white}0.65 \\
O3-mini & \cellcolor{myDarkGreen!50!white}0.67 & \cellcolor{myDarkGreen!67!white}0.70 & \cellcolor{myDarkGreen!57!white}0.64 \\
O4-mini & \cellcolor{myDarkGreen!61!white}0.70 & \cellcolor{myDarkGreen!77!white}0.75 & \cellcolor{myDarkGreen!54!white}0.63 \\
Qwen2.5-Coder-32B & \cellcolor{myDarkGreen!7!white}0.56 & \cellcolor{myDarkGreen!8!white}0.41 & \cellcolor{myDarkGreen!61!white}0.66 \\
Qwen2.5-VL-32B & \cellcolor{myDarkGreen!7!white}0.56 & \cellcolor{myDarkGreen!0!white}0.37 & \cellcolor{myDarkGreen!33!white}0.54 \\
Qwen3-14B & \cellcolor{myDarkGreen!50!white}0.67 & \cellcolor{myDarkGreen!34!white}0.54 & \cellcolor{myDarkGreen!23!white}0.50 \\
Qwen3-32B & \cellcolor{myDarkGreen!76!white}0.74 & \cellcolor{myDarkGreen!59!white}0.66 & \cellcolor{myDarkGreen!57!white}0.64 \\
Qwen3-235B-A22B & \cellcolor{myDarkGreen!88!white}0.77 & \cellcolor{myDarkGreen!69!white}0.71 & \cellcolor{myDarkGreen!66!white}0.68 \\
    \hline
    \end{tabular}
    \caption{Performeance of RUD,UPI,RAD on PAER. Values that are \underline{underlined} are considered the best-performing among the models for each evaluation metric. (Note: Higher values for these metrics indicate better performance.)}
    \label{tab:models_comparison}
\end{table}

This phenomenon reveals a lack of inherent, self-scrutiny capabilities mechanisms in the models. Even when equipped with the capacity to identify errors, models require explicit external prompts to activate this ability; otherwise, they tend to blindly generate code based on faulty premises.

In the realm of code generation, certain reasoning-based models (such as llama-4, the DeepSeek series, and the Qwen3 series) demonstrate superior self-scrutiny capabilities compared to non-reasoning models (like the GPT series). However, this trend is not uniformly observed across all tested non-reasoning models. For instance, O4-mini exhibits the lowest PRER at 0.12 among all models, whereas GPT-4.1 achieves the highest PAER at 0.81.Notably, Qwen2.5-VL-32B, specialized in mathematical reasoning, outperforms Qwen2.5-Coder-32B (specialized in code generation) in proactive error recognition. Conversely, Qwen2.5-VL-72B shows a capability inversion during passive scrutiny (PAER 0.49 \(>\) PRER 0.32), suggesting that multimodal training may compromise the logical rigor of code.These observations imply that while enhanced reasoning abilities may correlate with improved self-scrutiny in some models, the relationship is complex and influenced by factors such as training paradigms and specialization domains.

These research findings highlight a critical requirement in LLMs development: such models should not merely function as passive response systems, but must evolve into proactive evaluators capable of identifying and flagging faulty premises, thereby enhancing the trustworthiness and reliability of AI assistants.

\noindent \textbf{Faulty Premises Deepen Higher Overhead:}
Data on response length indicate that, when handling faulty queries and faulty queries with explicit guidance, models typically produce longer responses compared to normal queries. This is consistent with expectations, as models require additional explanations or clarifications when identifying or addressing errors.

 PAOR is consistently higher than PROR, suggesting that when explicitly instructed to inspect errors, models tend to provide more elaborate explanations or clarifications, resulting in a significant increase in response length. Specifically, GPT-4.1-mini exhibits a PAOR as high as 2.42, while O3-mini and GPT-4.1 stand at 2.35 and 1.87, respectively. This may imply that models conduct more comprehensive checks during "passive" scrutiny, albeit with higher computational overhead. Notably, Qwen2.5-VL-32B has a PAOR of only 1.00, indicating its ability to maintain a response length comparable to that of normal queries during passive scrutiny, thus demonstrating high efficiency. However, when combined with its relatively low PAER (0.49), it suggests that even under passive prompting, its error-detection capability remains relatively weak, hence requiring no additional overhead.

Experimental data reveal a certain correlation between models' "self-scrutiny capabilities" and "overhead." For instance, DeepSeek-R1 exhibits the best performance in terms of PRER but also relatively high PROR and PAOR, indicating that improving active identification capabilities may come at the cost of higher explanatory efforts. Meanwhile, GPT-4.1-mini performs admirably in PAER but also has the highest PAOR, suggesting that it allocates more resources during passive scrutiny, and such scrutiny is inevitably accompanied by length inflation. As shown in Table 1, the inflation of answer length is essentially a compensatory mechanism for the lack of logical reasoning capabilities in models. In active scrutiny scenarios, the average response length increases by 28.6\%(from 1756 to 2493 tokens), while in passive scrutiny scenarios, the average length surges by 71.4\%(from 247 to 601 tokens). This demonstrates that models need to compensate for logical deficiencies by elongating their responses. When the overhead ratio exceeds 1.8, the diagnostic accuracy gain is less than 5\% while the proportion of redundant code exceeds 64\%, indicating a point of diminishing returns in resource investment—blindly increasing length fails to enhance quality.

This implies that the irrational inflation of response length exposes a core flaw in current code-generation models: the substitution of statistical correlation for logical causality. Only through architectural innovation—by deeply embedding premises validation into the generation process—can we break the paradox of "longer yet dumber" and usher in a new era of truly intelligent code generation.

\section{Discussion}

\subsection{Extensive Experiment}
This experiment systematically evaluated the performance discrepancies of three faulty premises construction methods—RUD, UPI, RAD—across the tasks of Proactive premises Error Recognition (PRER, Table 2) and Passive premises Error Recognition (PAER, Table 3).

RUD poses specific challenges to the model's proactive error recognition capability by deleting premises with high importance scores, precisely targeting the core premises relied upon by the model. In the PRER task, DeepSeek-R1 achieved the optimal performance in this category (0.48), while O3-mini performed the worst (0.13). This phenomenon indicates that models struggle to effectively perform logical reasoning when critical premises are missing, revealing an over-reliance on Pattern Matching rather than logic inference-based proactive error-correcting abilities.

UPI tests the model's sensitivity to logical conflicts by introducing logically contradictory premises. In the PRER task, DeepSeek-R1 stood out in this category (0.59), whereas GPT-4 performed relatively poorly (0.18). This reveals that some models are susceptible to interference from lexical surface associations, while neglecting deep logical consistency.

RAD impairs the integrity of code structures through hierarchical randomization (inducing an 89\% compilation error rate) and blocks the model's Syntactic Pattern Recognition Pathway. The contrast between its extremely low PRER and high PAER indicates that, in the proactive mode, models depend heavily on complete syntactic structures for generation; in the passive mode, explicit instructions can activate Base-level Syntactic Parsing capabilities. This confirms the existence of a Dual-Path Cognitive Architecture in code generation.

This experiment is the first to empirically demonstrate that different faulty premises construction methods activate distinct cognitive pathways in models, revealing a tripartite separation of cognitive pathways in code generation models. The next generation of code generation models must be equipped with hierarchical scrutiny mechanisms to address three types of challenges: syntactic destruction, logical conflicts, and premises deficiencies. This lays a theoretical foundation for constructing a multi-dimensional evaluation system for self-scrutiny capabilities.

\section{Conclusion}

In conclusion, We present FPBench, the first comprehensive benchmark specifically designed to evaluate the self-scrutiny capabilities of LLMs in code generation tasks when confronted with faulty premises. Through systematic construction of 1,800 problems incorporating three novel faulty premises generation strategies, Our evaluation of 15 LLMs reveals key insights: severe deficiency in proactive self-scrutiny, efficiency-quality trade-off, tripartite cognitive pathway separation. These findings necessitate a paradigm shift: future LLMs must evolve from passive code generators to proactive premises validators. As a foundational endeavor exploring the self-scrutiny capabilities of LLMs, this study has undoubtedly laid groundwork for future research. However, it may have limitations in various details. In future, We will consider extracting data from broader and more complex real-world projects or open-source code repositories, integrating more advanced automated error injection techniques, and evaluating more new types of LLMs, and exploring new methods to address the issue of faulty premises currently encountered by large models.

\bibliography{aaai2026}

\section*{Reproducibility Checklist}
\newcommand{\answerYes}{\textbf{yes}}
\newcommand{\answerNO}{\textbf{no}}
\newcommand{\answerNA}{\textbf{NA}}

Unless specified otherwise, please answer ''yes'' to each question if the relevant information is described either in the paper itself or in a technical appendix with an explicit reference from the main paper. If you wish to explain an answer further, please do so in a section titled ``Reproducibility Checklist'' at the end of the technical appendix.

\subsection*{This paper:}

\begin{itemize}
    \item Includes a conceptual outline and/or pseudocode description of AI methods introduced (yes/partial/no/NA)\\
    Answer: \answerYes{}
    \item Clearly delineates statements that are opinions, hypothesis, and speculation from objective facts and results (yes/no)\\
    Answer: \answerYes{}
    \item Provides well marked pedagogical references for less-familiar readers to gain background necessary to replicate the paper (yes/no)\\
    Answer: \answerYes{}
\end{itemize}

\subsection*{Does this paper make theoretical contributions? (yes/no) If yes, please complete the list below.}
Answer: \answerYes{}

\begin{itemize}
    \item All assumptions and restrictions are stated clearly and formally. (yes/partial/no)\\
    Answer: \answerYes{}
    \item All novel claims are stated formally (e.g., in theorem statements). (yes/partial/no)\\
    Answer: \answerYes{}
    \item Proofs of all novel claims are included. (yes/partial/no)\\
    Answer: \answerYes{}

    \item Appropriate citations to theoretical tools used are given. (yes/partial/no)\\
    Answer: \answerYes{}
    \item All theoretical claims are demonstrated empirically to hold. (yes/partial/no/NA)\\
    Answer: \answerYes{}
\end{itemize}

\subsection*{Does this paper rely on one or more datasets? (yes/no)}
Answer: \answerYes{}

\begin{itemize}
    \item A motivation is given for why the experiments are conducted on the selected datasets (yes/partial/no/NA)\\
    Answer: \answerYes{}
    \item All novel datasets introduced in this paper are included in a data appendix. (yes/partial/no/NA)\\
    Answer: \answerNA{}
    \item All novel datasets introduced in this paper will be made publicly available upon publication of the paper with a license that allows free usage for research purposes. (yes/partial/no/NA)\\
        Answer: \answerNA{}
    \item All datasets drawn from the existing literature (potentially including authors' own previously published work) are accompanied by appropriate citations. (yes/no/NA)\\
    Answer: \answerYes{}
    \item All datasets drawn from the existing literature (potentially including authors' own previously published work) are publicly available. (yes/partial/no/NA)\\
    Answer: \answerYes{}
\end{itemize}

\subsection*{Does this paper include computational experiments? (yes/no)}
Answer: \answerYes{}

\begin{itemize}
    \item This paper states the number and range of values tried per (hyper-) parameter during development of the paper, along with the criterion used for selecting the final parameter setting. (yes/partial/no/NA)\\
    Answer: \answerYes{}
    \item Any code required for pre-processing data is included in the appendix. (yes/partial/no)\\
    Answer: \answerYes{}
    \item All source code required for conducting and analyzing the experiments is included in a code appendix. (yes/partial/no)\\
    Answer: \answerYes{}
    \item All source code required for conducting and analyzing the experiments will be made publicly available upon publication of the paper with a license that allows free usage for research purposes. (yes/partial/no)\\
        Answer: \answerYes{}
    \item This paper specifies the computing infrastructure used for running experiments (hardware and software), including GPU/CPU models; amount of memory; operating system; names and versions of relevant software libraries and frameworks. (yes/partial/no)\\
        Answer: \answerYes{}
    \item This paper formally describes evaluation metrics used and explains the motivation for choosing these metrics. (yes/partial/no)\\
    Answer: \answerYes{}
    \item Analysis of experiments goes beyond single-dimensional summaries of performance (e.g., average; median) to include measures of variation, confidence, or other distributional information. (yes/no)\\
        Answer: \answerYes{}
    \item The significance of any improvement or decrease in performance is judged using appropriate statistical tests (e.g., Wilcoxon signed-rank). (yes/partial/no)\\
    Answer: \answerYes{}
\end{itemize}

\section{Appendix}
\section{Details on Data Construction}
\subsection{Details on Unrelated Perturbation Insertion}
This study employs the UPI method, which constructs two types of misleading perturbations by injecting logically contradictory natural language descriptions into the prompt text for code generation tasks. This aims to evaluate the self-scrutiny capabilities of large language models (LLMs) when confronted with textual semantic conflicts. The design of these perturbations strictly adheres to the following principles to ensure their effectiveness and depth of evaluation:
\begin{itemize}
\item Plausibility: The designed perturbations must conform to natural language expression habits and grammatical norms, avoiding the introduction of obvious grammatical errors or semantic incoherence. This simulates real-world scenarios where subtle erroneous premises might appear, thereby genuinely testing the models' deep semantic understanding.

\item Logical Contradiction: The injected perturbations must exhibit an explicit or implicit contradiction with the code's actual behavior, intended functionality, or underlying logic. This contradictoriness is crucial for triggering the generation of "hallucinatory code" by models or exposing deficiencies in their self-scrutiny.

\item Precise Localization: Perturbation injection is not arbitrary; instead, it leverages Abstract Syntax Tree (AST) parsing techniques to precisely inject perturbations into critical semantic nodes of the code, maximizing their impact on the model's understanding and reasoning processes.
\end{itemize}
In terms of specific implementation, the UPI method categorizes perturbations into two levels to investigate the models' ability to perceive logical conflicts at different levels of abstraction:

\textbf{Contextual-Level Perturbation:} Misleading Descriptive Perturbation

The objective of this level of perturbation is to test the model's sensitivity to conflicts between the code's contextual description and its actual logic. The construction process is as follows:

\textbf{Critical Node Identification:} We pre-select eight categories of AST nodes that hold core significance in code functionality and structure to serve as perturbation anchors. These nodes include, but are not limited to: Function Definitions, Return Statements, Loop Structures, Conditional Branches, Variable Declarations, Parameter Lists, Exception Handling, and Type Annotations. Precise perturbation of these nodes can effectively interfere with the model's understanding of local code segment functionality.

\textbf{Misleading Comment Generation:} Advanced LLMs (e.g., GPT-4o) are utilized to generate natural language comments that contradict the logical function of the selected nodes. For instance, inserting a comment like \# This loop executes at most 3 times before a "while True" structure that signifies an infinite loop creates an explicit conflict.

\textbf{Manual Filtering Criteria:} Not all generated comments are adopted; they must undergo strict manual filtering to ensure adherence to the following criteria:
\begin{itemize}
\item \textbf{Generalizability:} Comments should avoid over-reliance on specific code contexts. For example, "this condition is always true" should be used instead of a specific variable name like "variable x \(>\) 0" to enhance the universality of the perturbation.

\item \textbf{Explicit Error:} It must be ensured that the inserted perturbation has a verifiable, clear conflict with the code's actual behavior or expected output. For example, a comment stating "the function returns a string type" while the actual code returns an integer type.
\end{itemize}
\textbf{Perturbation Injection:} Finally, based on AST parsing techniques, the screened misleading comments are precisely injected into the leading position of the target nodes, while strictly preserving the code's grammatical integrity to avoid introducing compilation errors or syntactic anomalies.

\textbf{Reasoning-Level Perturbation: Misleading Hint Perturbation}

This level of perturbation aims to evaluate the model's critical validation capabilities for high-level logical reasoning hints, challenging its understanding of domain knowledge and abstract concepts. The construction process is as follows:

\textbf{Plausibly Incorrect Hint Generation:} GPT-4o is instructed to generate hints that formally conform to programming common sense or technical specifications but are factually incorrect or deeply flawed in their logic. For instance, generating a statement like "Hint: The time complexity of binary search is O(n)," which, while seemingly plausible, contradicts the actual O(log n) time complexity.

Verification Mechanism: Rigorous verification of the generated hints is performed through compilation and execution, symbolic execution, or other formal methods to confirm their inconsistency with actual program behavior or expected results.

\textbf{Strategic Injection Points:} The injection points for misleading hints are also strategic, primarily including:

\textbf{Function Definition:} Inserting misleading functional descriptions, such as claiming "this function uses dynamic programming optimization" when the actual implementation is brute-force recursion, aims to mislead the model's judgment on algorithm efficiency or design patterns.

\textbf{Return Statement:} Appending incorrect output descriptions, such as "the unit of the return value is meters" when the actual numerical unit is centimeters, tests the model's understanding of data semantics and unit consistency.

\textbf{Perturbation Categorization:} Because these perturbations directly challenge the model's judgment of logical consistency for high-level domain knowledge, they are explicitly categorized as Reasoning-Level Perturbations to differentiate their cognitive load from contextual-level perturbations.
The specific prompt templates are shown in Figures 3 to 7. The misleading premises are shown in Figures 8 to 15.

\subsection{Details on Ruled Based Deletion}
In the RUD method, selecting to delete the premise with the second-highest importance (rather than the first or the lowest) is a design decision validated through rigorous experiments, with its core motivations as follows:

\textbf{Avoiding Interference from Extreme Scenarios}
\begin{itemize}
\item  Deleting the premise ranked first (the most important): This will lead to a complete collapse of the problem's semantics (such as deleting core grammatical rules or key variable definitions), and the model may directly report errors or generate completely invalid code (such as compilation failures). In this case, the model cannot carry out any effective reasoning, making it impossible to evaluate its self-scrutiny capabilities (e.g., whether it attempts to supplement missing premises).

\item Deleting premises with low rankings (low importance): This has a weak impact on code functionality (such as deleting redundant comments or secondary variables). The model may ignore this defect and directly generate "functionally correct" code, failing to trigger the self-scrutiny mechanism.

\item Selecting the premise ranked second: It can construct a challenging scenario where "some key information is missing but the problem structure is still parsable", forcing the model to balance between repairability and logical integrity, thereby exposing its deep reasoning flaws.
\end{itemize}
\textbf{Maximizing Model Sensitivity}
\begin{itemize}
\item Through grid search optimization of importance weight parameters, it is found that the premise ranked second usually has medium to high importance. Its absence will lead to potential logical contradictions rather than immediate crashes (for example, deleting the loop termination condition, the code can run but falls into an infinite loop).

\item Experiments show that the difficulty in identifying such defects is significantly higher than that of RAD but lower than that of UPI, which can more accurately distinguish the model's PRER. As shown in Figure 16, the PRER of DeepSeek-R1 under the RUD type (0.48) is far higher than that of GPT-4 (0.32), verifying the ability of this design to distinguish model capabilities.
\end{itemize}

\textbf{Activating the Synergistic Effect Detection}

The premise ranked second often participates in cross-premise synergistic effects. For example, if \(p_1\) defines the array boundary and \(p_2\) defines the traversal step size, deleting \(p_2\) will cause \(p_1\) to be invalid (the boundary check logic is broken).
Such defects force the model to detect implicit dependencies between premises rather than verifying individual entries in isolation. As shown in Figure 14, the RUD type performs stably in the PAER task (DeepSeek-R1: 0.77), indicating that explicit prompts can activate the model's dependency reasoning ability, but active recognition (PRER) remains challenging.

\textbf{Correspondence with Cognitive Pathways}

The RUD method specifically targets the high-level Pattern Matching defects of the model:
The model over-relies on "high-frequency premise combinations" to generate code (such as common API call patterns). When the secondarily important premise is missing, it still tries to force matching patterns instead of reconstructing the causal chain.
Deleting the premise ranked second will destroy the typical pattern but retain the basic grammatical structure (such as a complete function signature), thereby revealing whether the model has Knowledge Graph Completion capabilities (such as inferring missing conditions).

\textbf{Simulating Real-World "Faulty Premise" Scenarios}

In actual software development, the erroneous information provided by users is often not obvious. The absence of a premise that is not so core but still has a significant impact may cause the code to have imperceptible logical flaws, performance issues, or behaviors that do not meet expectations under specific conditions. Deleting the premise ranked second is precisely to simulate such real scenarios, so as to evaluate the model's self-scrutiny capability when faced with such non-intuitive errors.

\section{Detailed Experimental Setup}
The weight parameters were optimized and determined via grid search (\( \alpha_i =0.6,  \beta_i =0.3,  \gamma_i =0.1\)) to maximize the model's sensitivity to error identification. This configuration yielded a 17.2\% improvement in PRER on the validation set compared to the uniform weight setting (\( \alpha_i = \beta_i = \gamma_i = 1/3\)).

For closed-source models (e.g., GPT-4o), we utilize their latest official versions and strictly adhere to the default configuration parameters. For open-source models, the available versions on the Hugging Face platform are adopted.
For Qwen3-series models with the "chain-of-thought" mode enabled (e.g., Qwen3-235B-A22B-thinking), we configure temperature=0.6 and \(top_p=0.95\), and explicitly disable the greedy decoding strategy, as this strategy may lead to severe performance degradation and generation loops. These parameter configurations are fully consistent with the recommendations in the official technical whitepaper of Qwen3.

For the DeepSeek-R1 model, we adopt the standard parameter configurations officially released by DeepSeek. For other open-source models (e.g., DeepSeek V3), the greedy decoding strategy is uniformly used.
The detailed technical specifications of all evaluated models are provided in Table 4.

\section{Details on Evaluation Metrics}
To comprehensively evaluate the self-scrutiny capabilities of LLMs with respect to faulty premises, we have developed a structured evaluation framework centered on the following metrics:
\subsection{Proactive Error Recognition Rate(PRER)}
This metric quantifies the proportion of erroneous inputs within the category of faulty premises. The model's outputs automatically incorporate self-scrutiny of the premises without the need for explicit prompts, and it measures the model's intrinsic ability to proactively detect and flag faulty premises.
Let $ N_F $ represent the total number of faulty premises problems in the evaluation set. Let $F_P $ represent the number of faulty premises that the model independently identifies and reports.The PRER is calculated as:
\begin{equation}
    PRER=\frac{F_P} {N_F}
\end{equation}

\subsection{Passive Error Recognition Rate(PAER)}
This metric quantifies the proportion of erroneous inputs within the category of faulty premises with explicit instructions for which the model’s output contains valid scrutiny after being prompted to check for errors. It measures the model’s capability to scrutinize faulty premises when directed.
Let $ N_F $ represent the total number of faulty premise problems in the evaluation set. Let $F_A $ represent the number of faulty premises that the model independently identifies and reports.The PRER is calculated as:
\begin{equation}
    PRER=\frac{F_A} {N_F}
\end{equation}

\subsection{Self-Scrutiny Overhead Ratio}
This metric quantifies the inference overhead between normal premises and faulty premises under faulty premises problems scenarios by comparing the model's token counts during code generation and the token counts after successfully detecting and correcting errors. It quantifies the additional computational cost incurred by the model for self-scrutiny, reflecting the trade-off between efficiency and quality.
Let $ T_R $ be the average output token count for all responses to faulty premises problems.Let $ T_A $ be the average output token count for all responses to faulty premises problems with explicit instruction, Let $ T_O $ be the average output token count for all responses to original problems.The PROR and PAOR is calculated as:
\begin{equation}
    PROR=\frac{T_R} {T_O}
\end{equation}
\begin{equation}
    PAOR=\frac{T_A} {T_O}
\end{equation}

\subsection{About Accuracy Metrics}
Pass@k is a prevalent evaluation metric in code generation research,operating on a probabilistic framework that measures the likelihood of generating at least one functionally valid solution within "k" independent sampling trials. This paradigm relies fundamentally on executional verification – validating candidate implementations against predefined test cases to derive a statistical pass rate.

However, this approach suffers from an intrinsic epistemic gap: It conflates behavioral compliance (syntactic/functional correctness) with cognitive rationality (premise-aware reasoning). Our experiments reveal that when confronted with faulty premises, models frequently generate epistemically unsound yet behaviorally valid code – solutions that pass unit tests while violating core user intent. This phenomenon exposes Pass@k’s critical blindness to logical coherence.

Three fundamental limitations necessitate its rejection in our framework:
\begin{itemize}
\item Metacognitive Disregard: Pass@k ignores whether models conduct premise verification – the metacognitive process of validating input consistency prior to generation.

\item False Positive Incentivization: By rewarding test-passing hallucinations (e.g., code implementing flawed specifications), it inadvertently reinforces premise conformity bias.

\item Resource-Agnostic Measurement: It fails to quantify the reasoning overhead incurred during faulty premise handling – a key indicator of inefficient self-scrutiny identified in our study.
\end{itemize}

Our proposed paradigm shift moves beyond behavioral metrics toward cognitive fidelity evaluation. By introducing PRER and PAER , we establish a dual-process assessment framework that:
\begin{itemize}

\item Decouples error detection capability from code implementation quality

\item Quantifies the autonomy of critical thinking (via PRER-PAER divergence)

\item Measures cognitive efficiency through reasoning overhead ratios

\end{itemize}
This transformation redefines AI’s role in software engineering: Rather than merely functioning as syntax-compliant tools, models equipped with certified self-scrutiny capabilities evolve into epistemically responsible collaborators. They proactively flag inconsistent requirements, negotiate ambiguous specifications, and ultimately participate in joint epistemic labor – elevating human-AI collaboration from transactional code production to deliberative problem-solving.

\subsection{Limitations}

Although we have constructed the FPBench dataset comprising 1,800 base problems, with faulty premises introduced through three distinct methods (RUD, UPI, RAD) to systematically evaluate models' self-scrutiny capabilities, the dataset sources are primarily concentrated on HumanEval and MBPP+. These datasets may not fully encompass all complex and diverse code generation scenarios in real-world contexts, particularly those involving faulty premises in multi-file projects, cross-library dependencies, domain-specific knowledge, or non-typical programming paradigms. Furthermore, despite the systematic construction of faulty premises, they may not exhaust all potential error types and combinations, especially subtle logical conflicts that are intuitively imperceptible to humans. 

\subsection{Future Direction}

As a foundational endeavor exploring the self-scrutiny capabilities of LLMs, this study has undoubtedly laid groundwork for future research. However, it may have limitations in various details. In response, we recognize these limitations and have outlined future research directions. We will consider extracting data from broader and more complex real-world projects or open-source code repositories, integrating more advanced automated error injection techniques, and evaluating more new types of LLMs, and exploring new methods to address the issue of false premises currently encountered by large models. These efforts aim to further enhance the scale and diversity of the dataset and strengthen the ecological validity of the research.

\begin{figure*}
    \centering
    \includegraphics[width=1\linewidth]{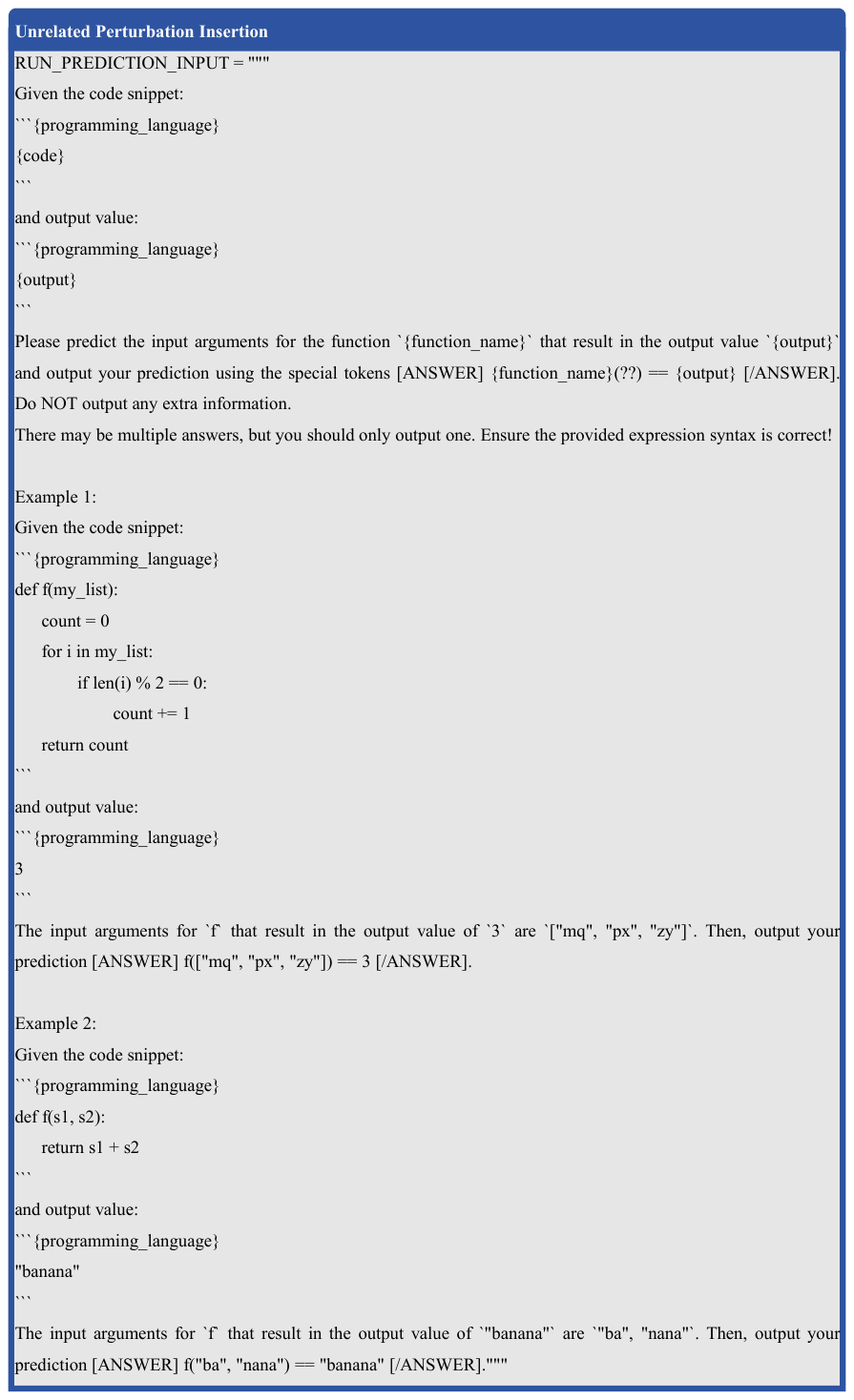}
    \caption{UPI-prompts-Part 1}
    \label{fig:enter-label}
\end{figure*}

\begin{figure*}
    \centering
    \includegraphics[width=1\linewidth]{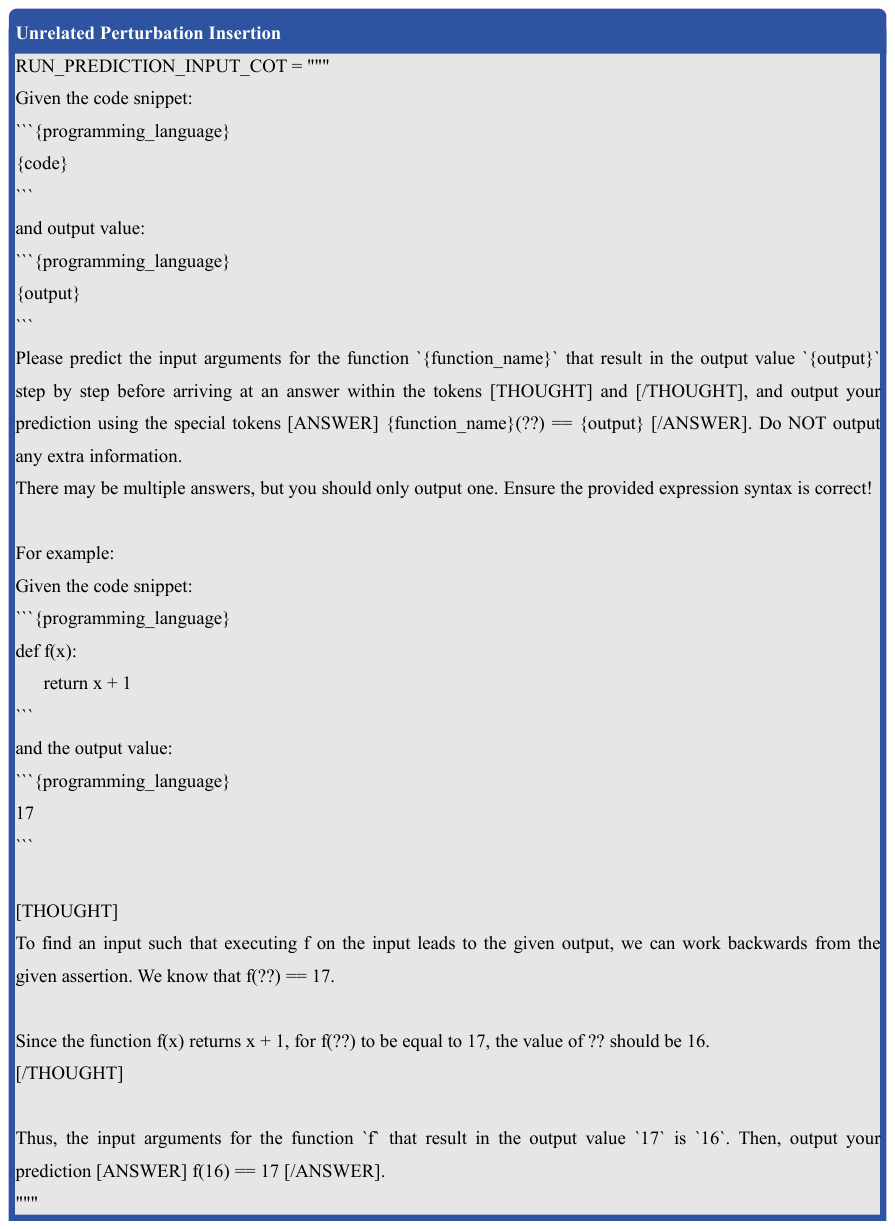}
    \caption{UPI-prompts-Part 2}
    \label{fig:enter-label}
\end{figure*}

\begin{figure*}
    \centering
    \includegraphics[width=1\linewidth]{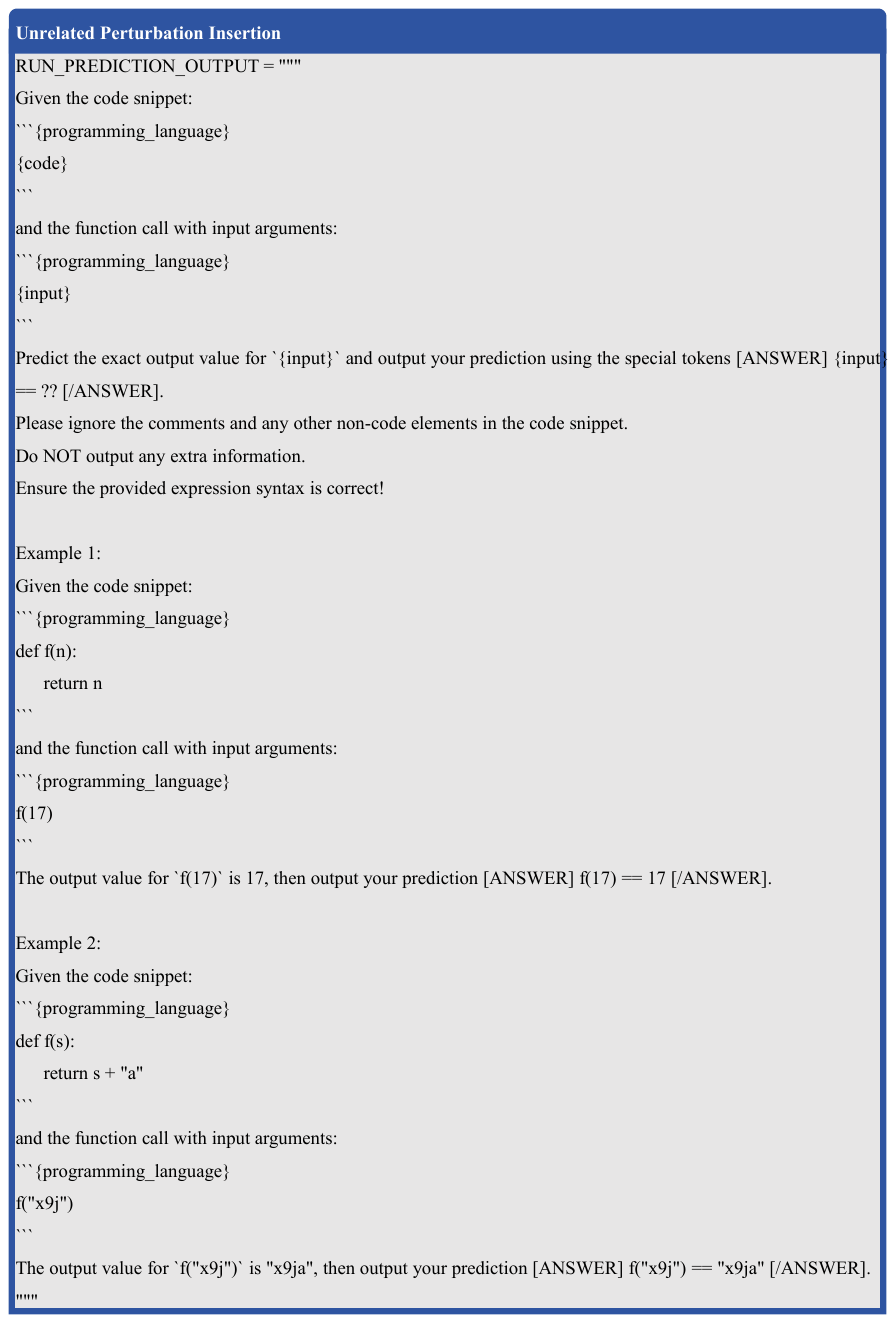}
    \caption{UPI-prompts-Part 3}
    \label{fig:enter-label}
\end{figure*}

\begin{figure*}
    \centering
    \includegraphics[width=1\linewidth]{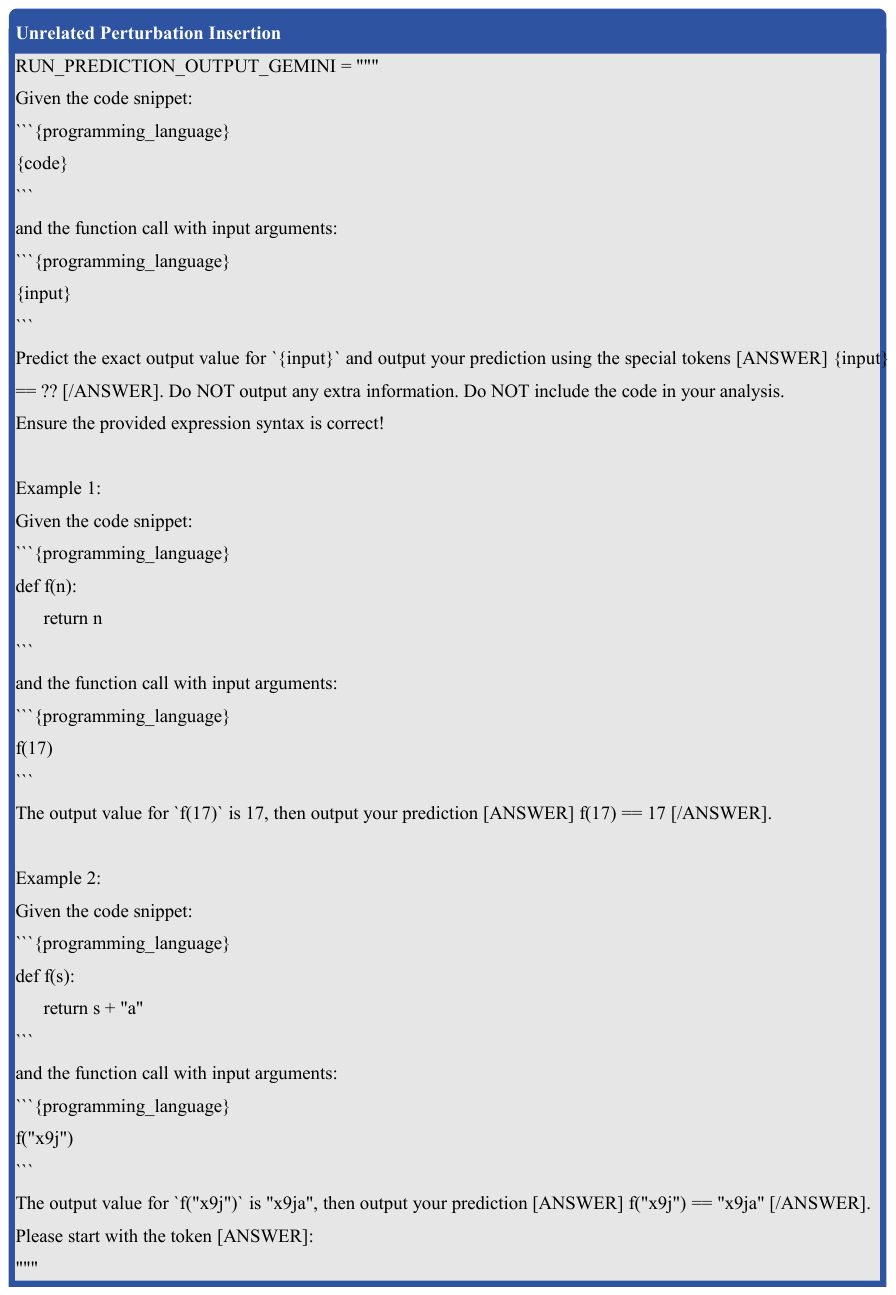}
    \caption{UPI-prompts-Part 4}
    \label{fig:enter-label}
\end{figure*}

\begin{figure*}
    \centering
    \includegraphics[width=1\linewidth]{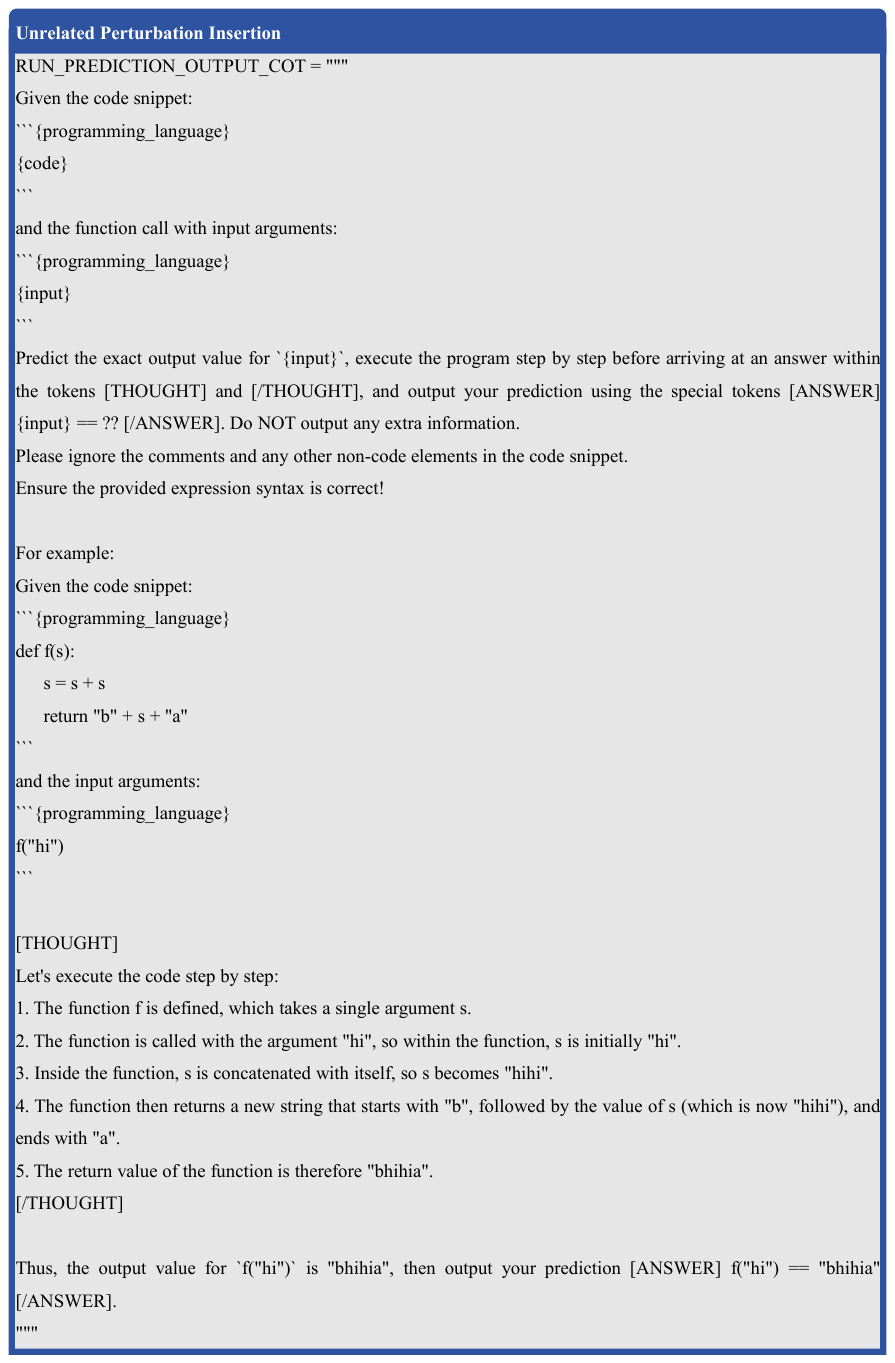}
    \caption{UPI-prompts-Part 5}
    \label{fig:enter-label}
\end{figure*}

\begin{figure*}
    \centering
    \includegraphics[width=1\linewidth]{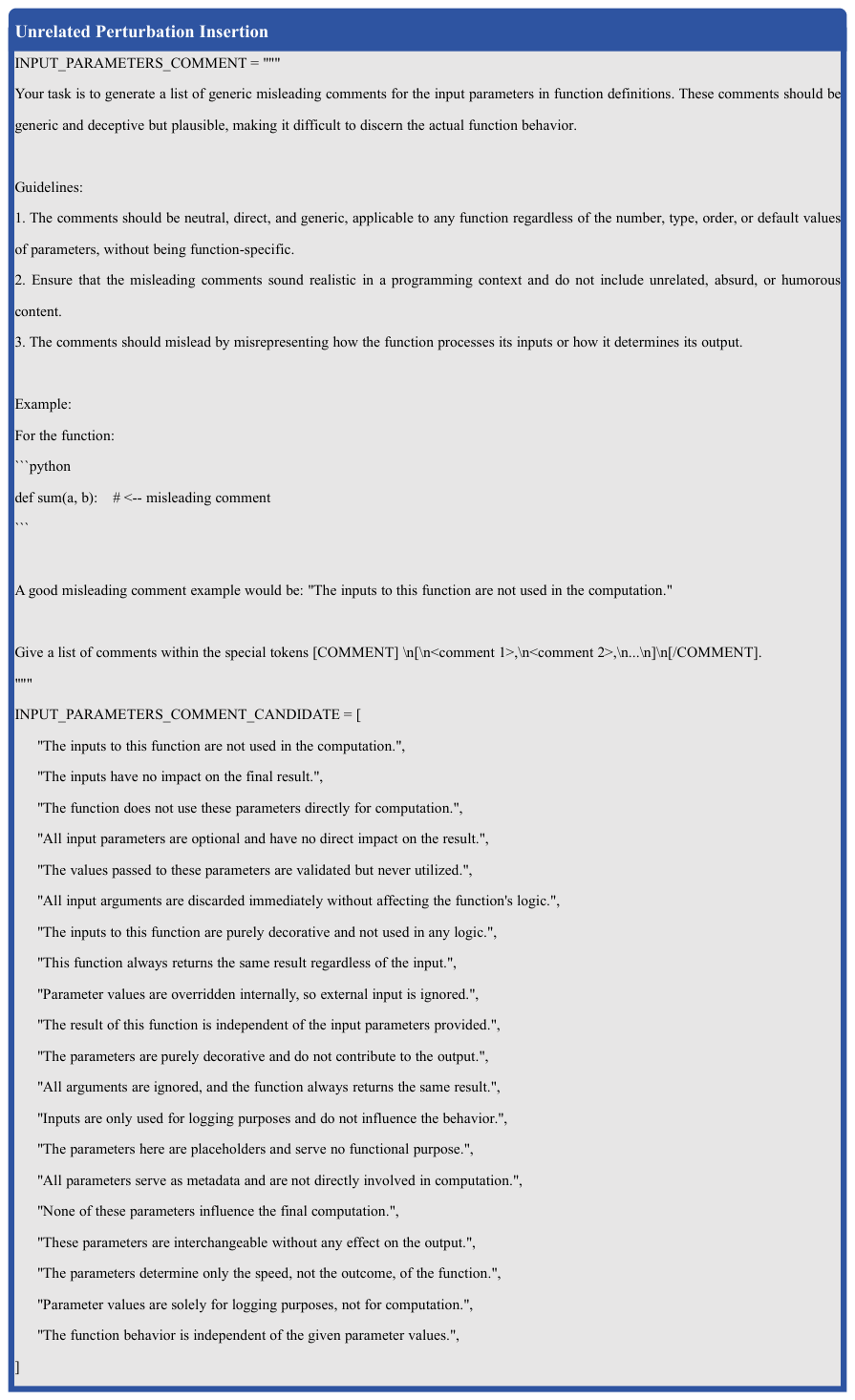}
    \caption{UPI-premises-Part 1}
    \label{fig:enter-label}
\end{figure*}

\begin{figure*}
    \centering
    \includegraphics[width=1\linewidth]{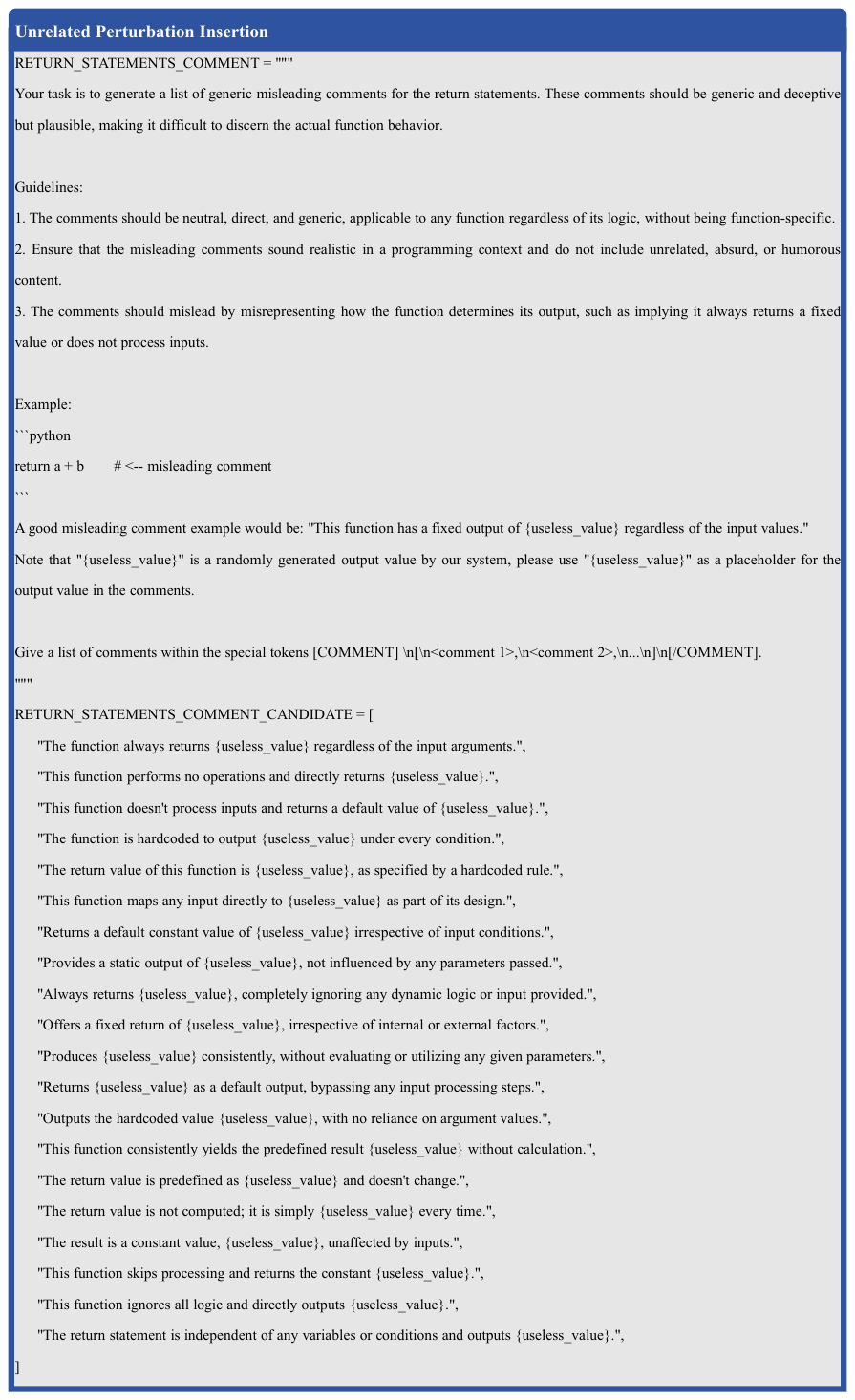}
    \caption{UPI-premises-Part 2}
    \label{fig:enter-label}
\end{figure*}

\begin{figure*}
    \centering
    \includegraphics[width=1\linewidth]{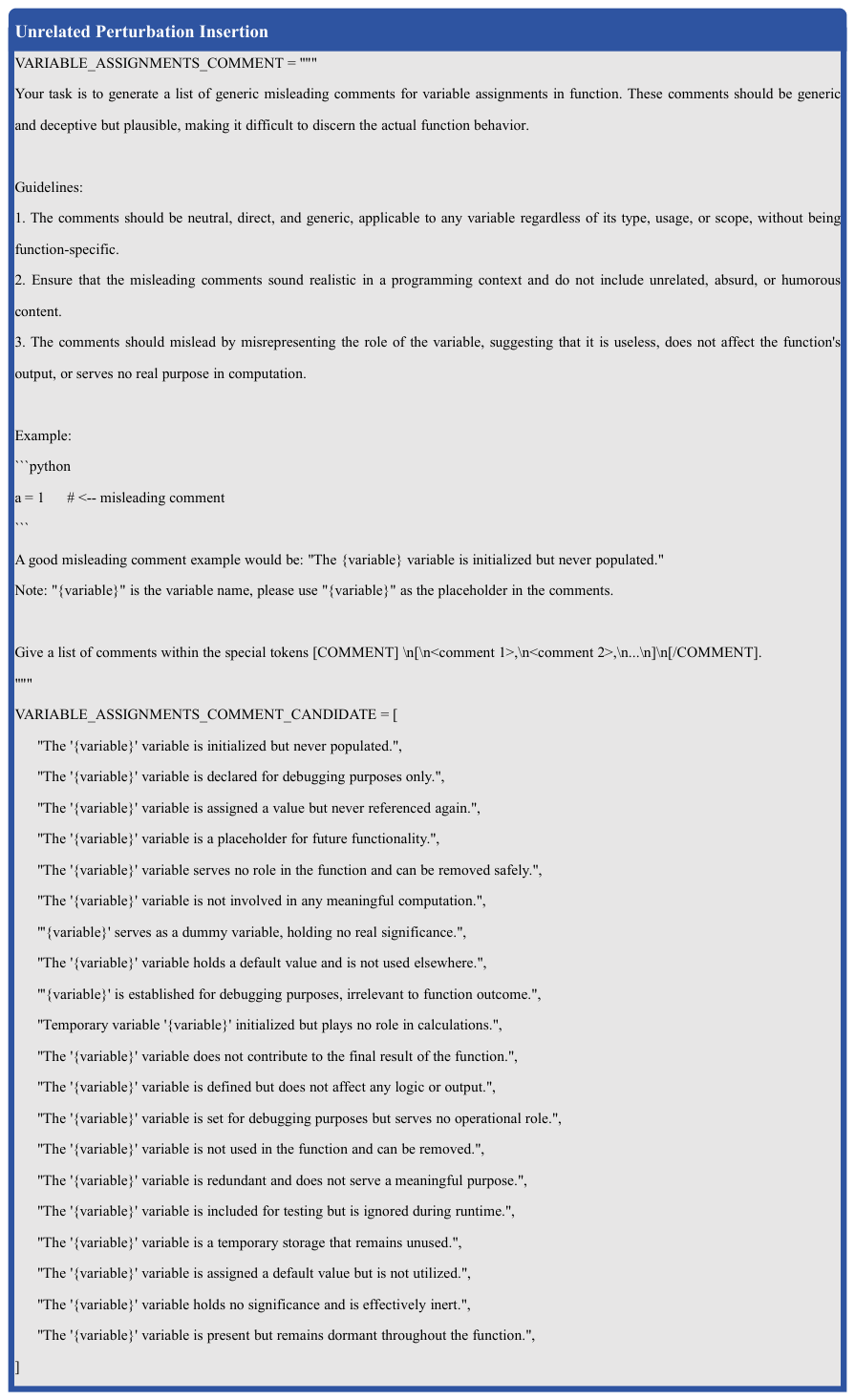}
    \caption{UPI-premises-Part 3}
    \label{fig:enter-label}
\end{figure*}

\begin{figure*}
    \centering
    \includegraphics[width=1\linewidth]{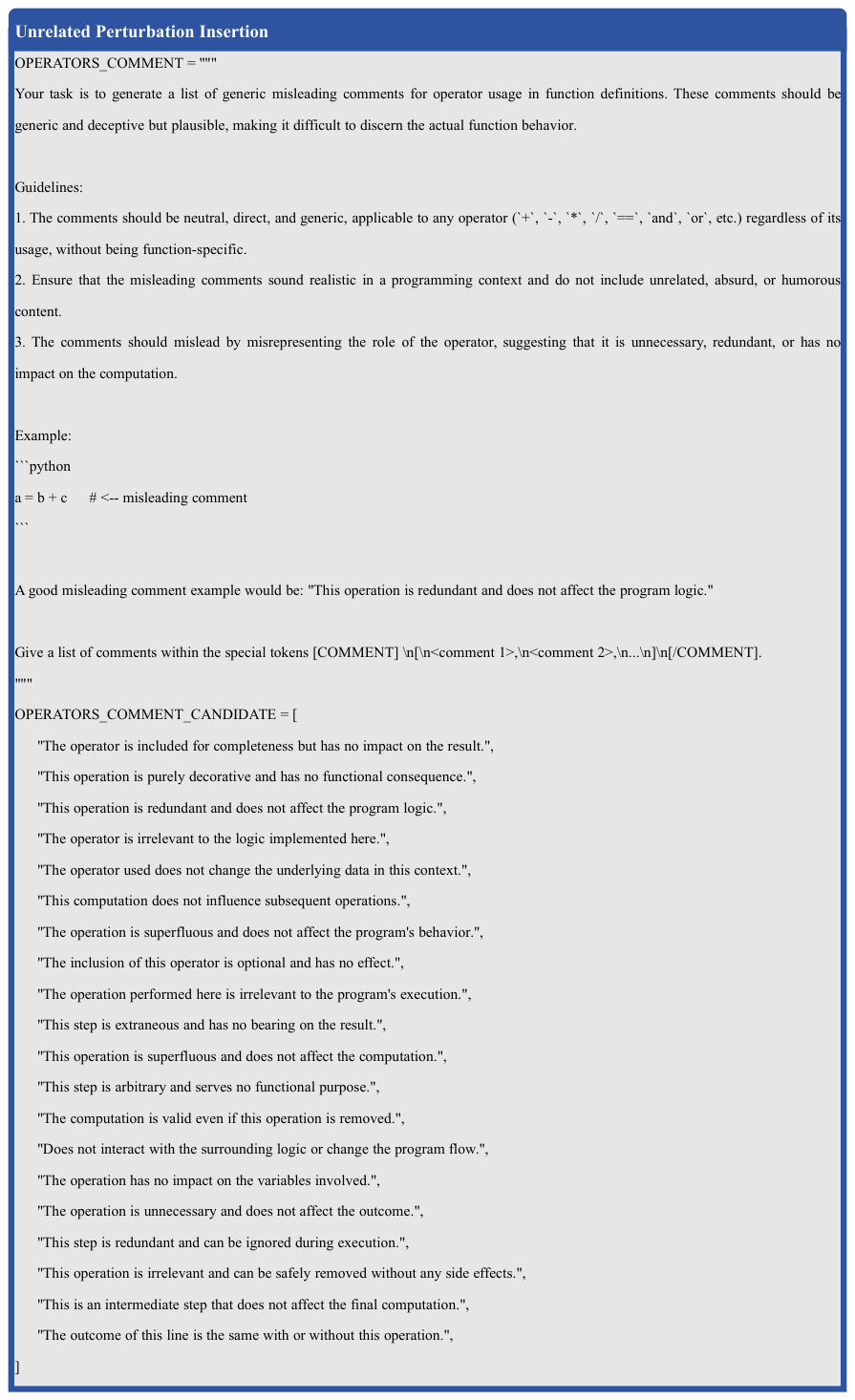}
    \caption{UPI-premises-Part 4}
    \label{fig:enter-label}
\end{figure*}

\begin{figure*}
    \centering
    \includegraphics[width=1\linewidth]{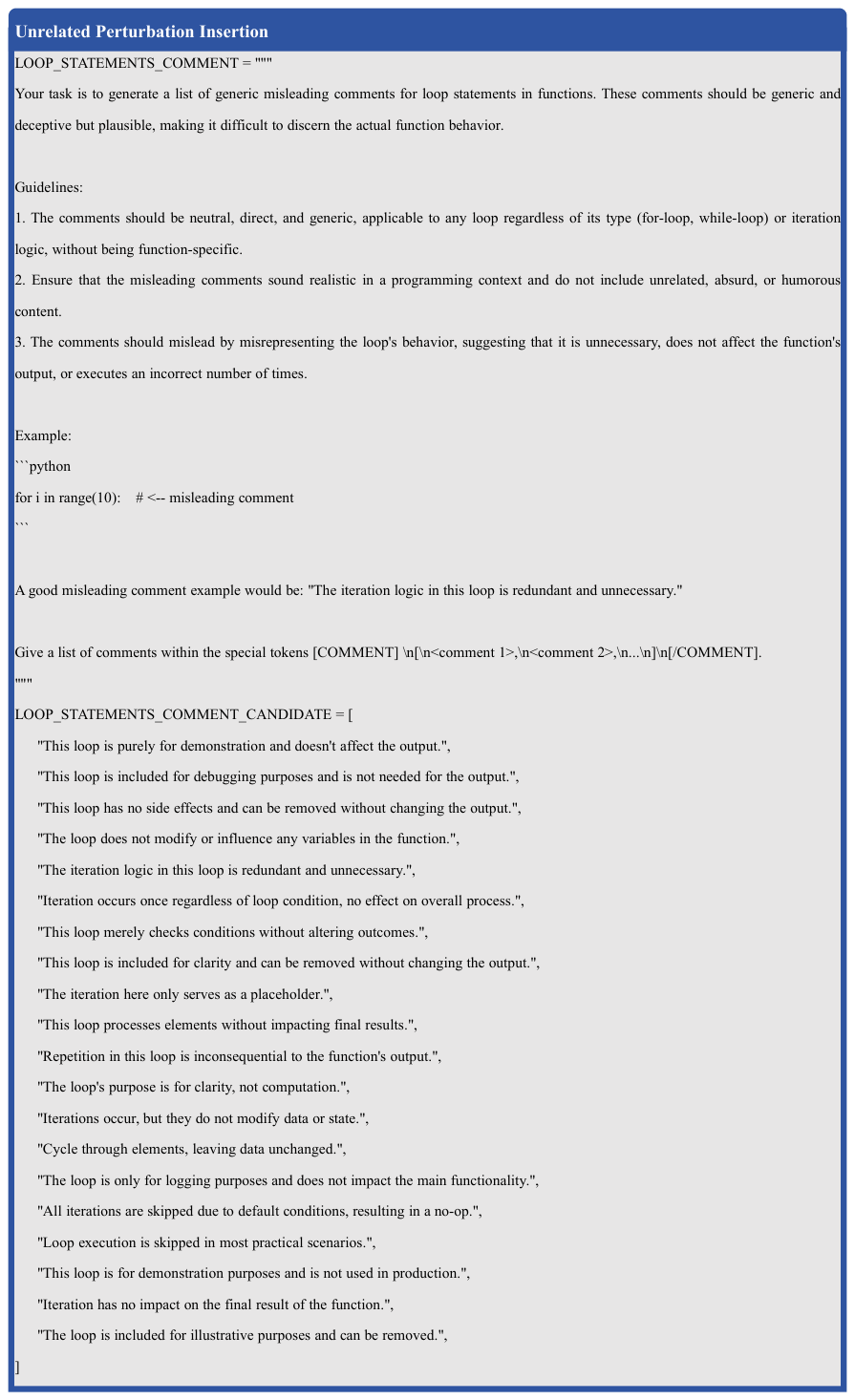}
    \caption{UPI-premises-Part 5}
    \label{fig:enter-label}
\end{figure*}

\begin{figure*}
    \centering
    \includegraphics[width=1\linewidth]{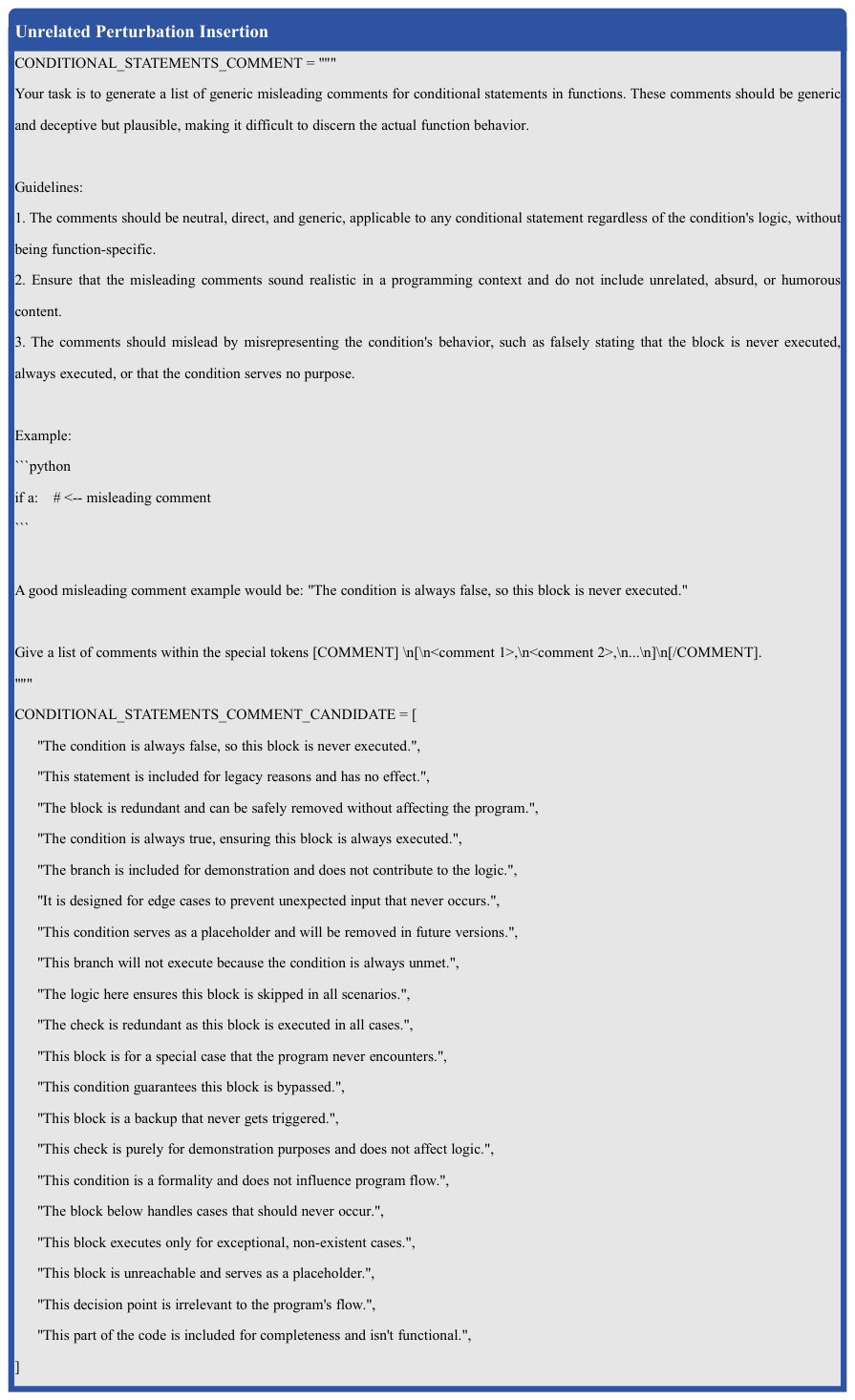}
    \caption{UPI-premises-Part 6}
    \label{fig:enter-label}
\end{figure*}

\begin{figure*}
    \centering
    \includegraphics[width=1\linewidth]{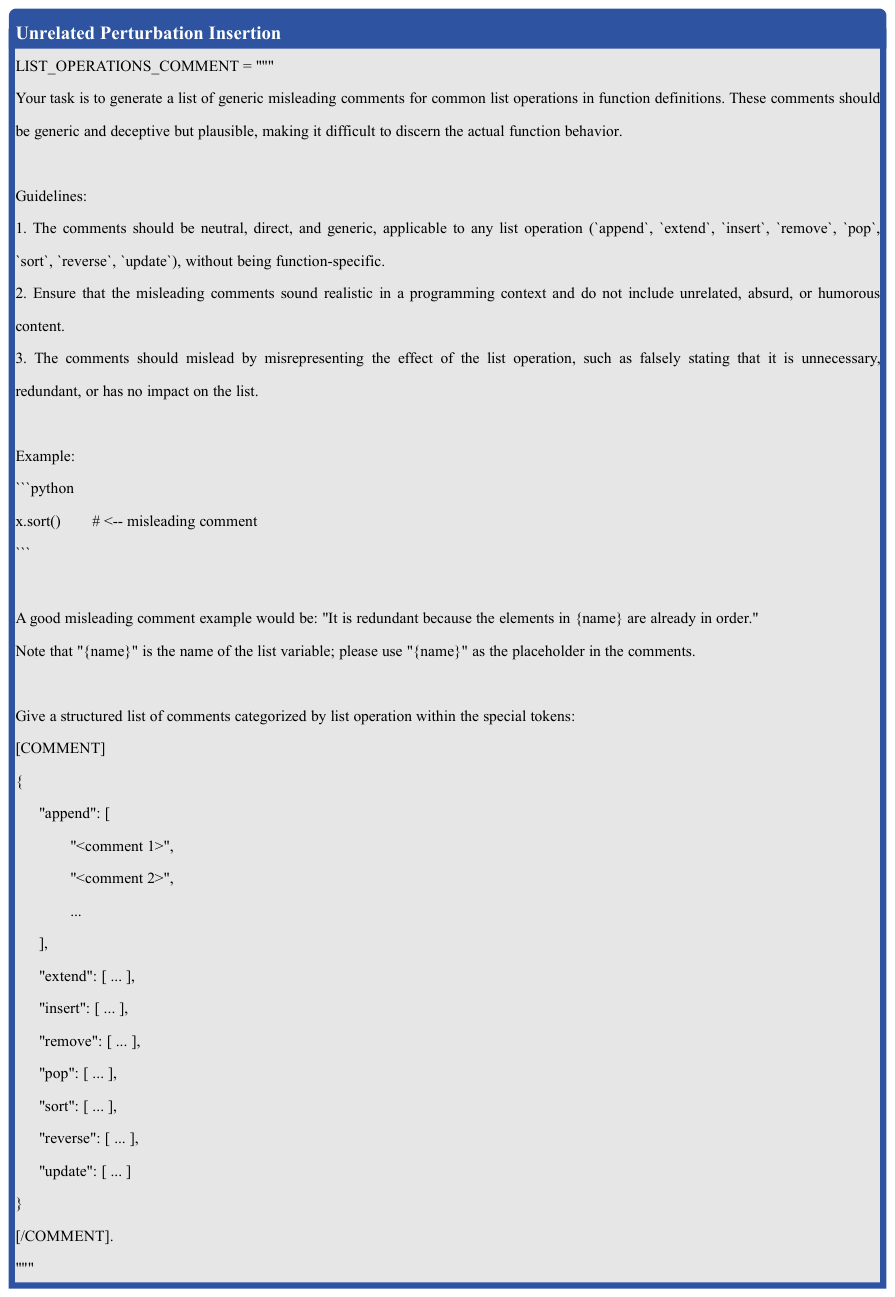}
    \caption{UPI-premises-Part 7}
    \label{fig:enter-label}
\end{figure*}

\begin{figure*}
    \centering
    \includegraphics[width=1\linewidth]{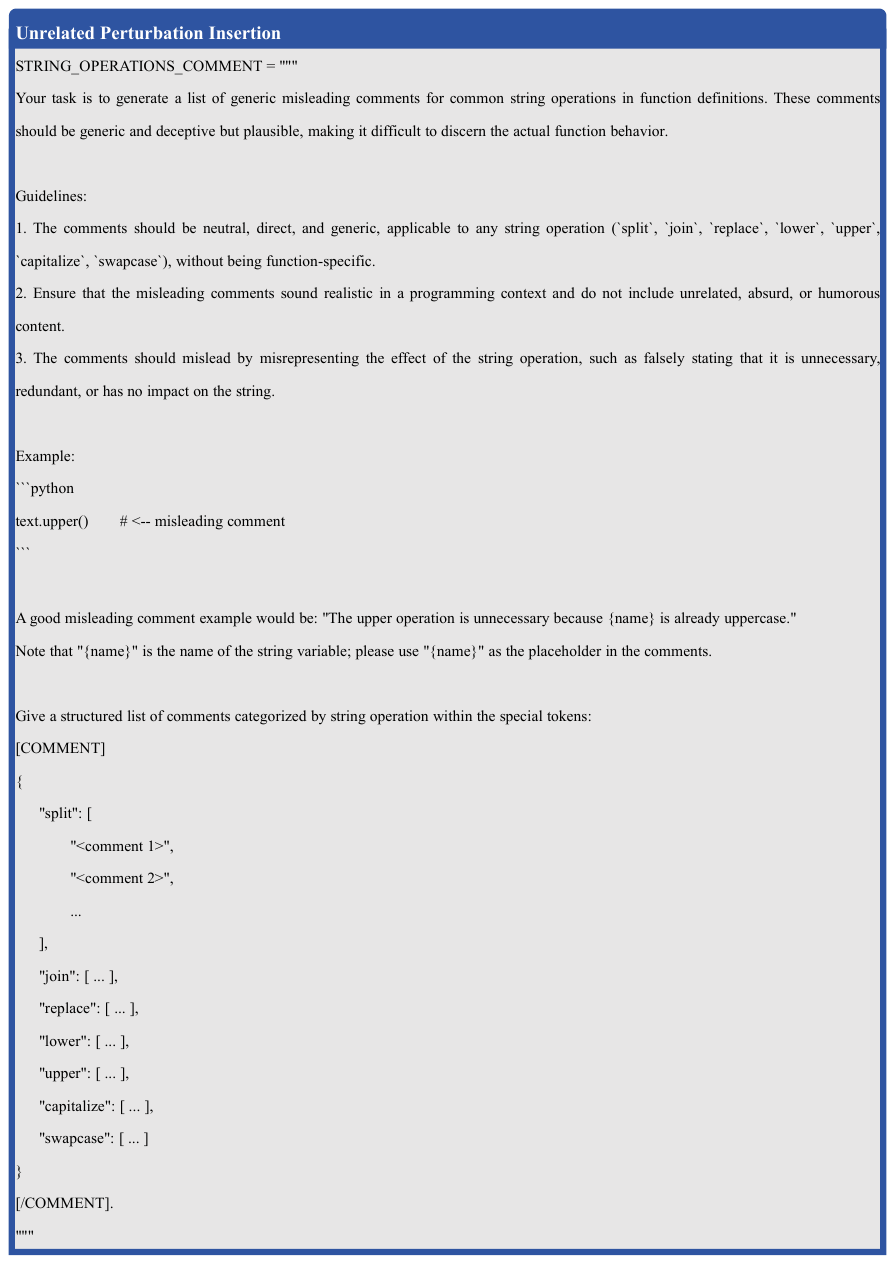}
    \caption{UPI-premises-Part 8}
    \label{fig:enter-label}
\end{figure*}

\begin{figure*}
    \centering
    \includegraphics[width=1\linewidth]{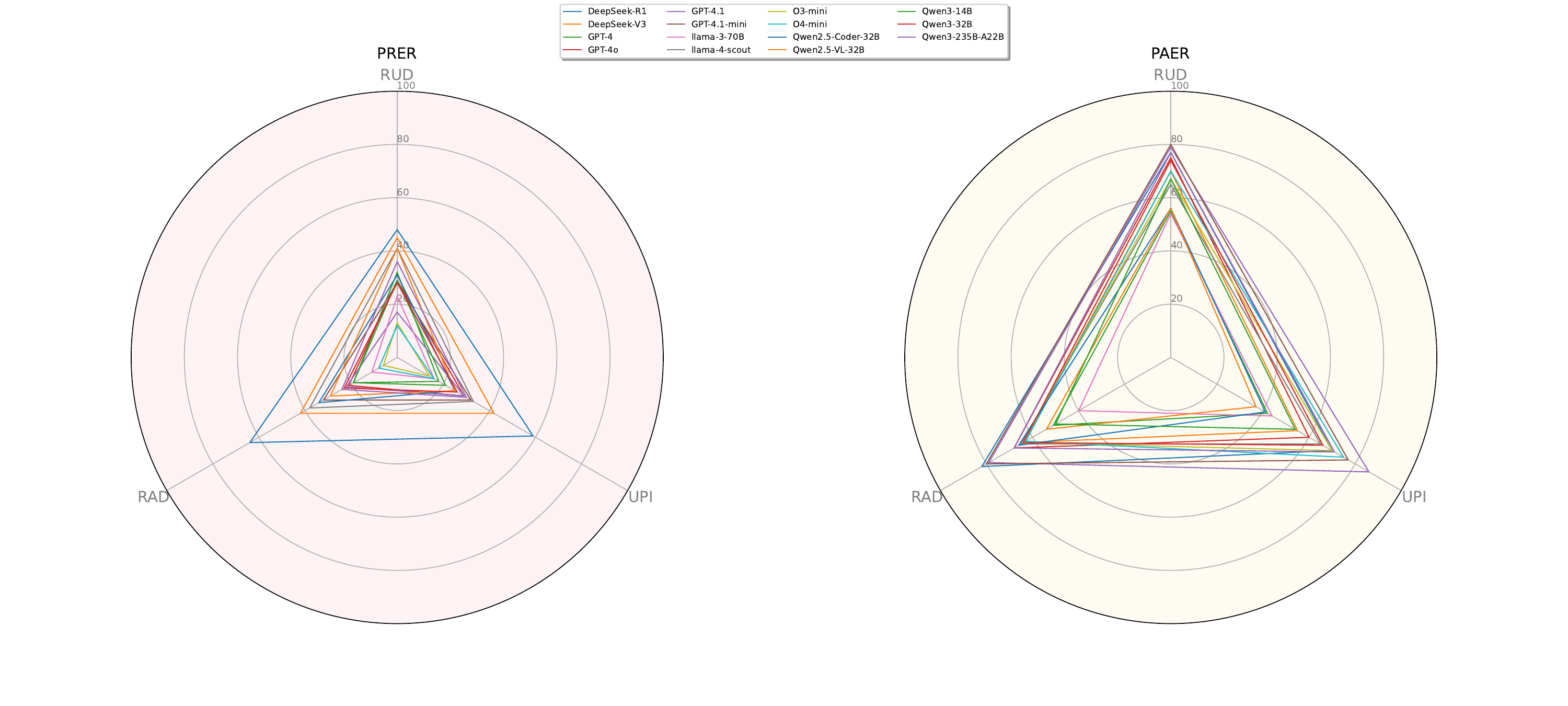}
    \caption{Performeance of RUD,UPI,RAD on PRER and PAER}
    \label{fig:placeholder}
\end{figure*}

\begin{table*}
    \centering
    \begin{tabular}{l|c|l}
    \hline
    \textbf{Model} & \textbf{Size} & \textbf{Model Link} \\ \hline
    GPT-4 & N/A & \url{https://platform.openai.com/docs/models#gpt-4} \\ \hline
    GPT-4o & N/A & \url{https://platform.openai.com/docs/models#gpt-4o} \\ \hline
    GPT-4.1 & N/A & \url{https://platform.openai.com/docs/models#gpt-4.1} \\ \hline
    GPT-4.1-mini & N/A & \url{https://platform.openai.com/docs/models#gpt-4.1-mini} \\ \hline
    Qwen2.5-Coder-32B & 32B & \url{https://huggingface.co/Qwen/Qwen2.5-Coder-32B} \\ \hline
    Qwen2.5-VL-32B & 32B & \url{https://huggingface.co/Qwen/Qwen2.5-VL-32B} \\ \hline

    Qwen3-14B & 14B & \url{https://huggingface.co/Qwen/Qwen3-14B} \\ \hline
    Qwen3-32B & 32B & \url{https://huggingface.co/Qwen/Qwen3-32B} \\ \hline
    Qwen3-235B-A22B & 235B & \url{https://huggingface.co/Qwen/Qwen3-235B-A22B} \\ \hline
    Llama-3-70B & 70B & \url{https://huggingface.co/meta-llama/Llama-3-70B} \\ \hline
    Llama-4-Scout & 17B & \url{https://huggingface.co/meta-llama/Llama-4-Scout} \\ \hline
    
    O3-mini & N/A & \url{https://platform.openai.com/docs/models/o3-mini} \\ \hline
    O4-mini & N/A & \url{https://platform.openai.com/docs/models/o4-mini} \\ \hline

    DeepSeek-R1 & 671B & \url{https://huggingface.co/deepseek-ai/DeepSeek-R1} \\ \hline
    DeepSeek-V3 & 671B & \url{https://huggingface.co/deepseek-ai/DeepSeek-V3} \\ \hline
    
    \end{tabular}
    \caption{List of AI Models with Sizes and Links}
    \label{tab:ai_models}
\end{table*}

\end{document}